\documentclass[10pt,twocolumn,letterpaper]{article}
\usepackage[pagenumbers]{cvpr}
\usepackage{graphicx}
\usepackage{amsmath,amssymb,amsthm}
\usepackage{booktabs,array,tabularx,multirow}
\usepackage{pifont}
\DeclareMathOperator*{\argmin}{arg\,min}
\newtheorem{proposition}{Proposition}
\newcommand{\cmark}{\ding{51}}
\newcommand{\xmark}{\ding{55}}
\newcommand{\UD}{U_{\mathrm D}}
\newcommand{\dAP}{d_{\mathrm{AP}}}
\newcommand{\dM}{d_{\mathrm{MSE}}}

\definecolor{cvprblue}{rgb}{0.21,0.49,0.74}
\usepackage[pagebackref,breaklinks,colorlinks,allcolors=cvprblue]{hyperref}
\hypersetup{
  pdftitle={Risk-Triggered Search: Behavior-Preserving Safe Control with Frozen Visual World Models},
  pdfauthor={Xinyu Deng, Bohao Lan, Kunpeng Li, Mingkai Yuan, Weihang Pan, Binbin Lin},
  pdfsubject={Visual world models and safe control},
  pdfkeywords={visual world models, safe control, risk-triggered search, JEPA}
}
\title{Risk-Triggered Search: Behavior-Preserving Safe Control with Frozen Visual World Models}
\author{
Xinyu Deng\quad Bohao Lan\quad Kunpeng Li\\
Mingkai Yuan\quad Weihang Pan\thanks{Corresponding author: \href{mailto:panweihang@zju.edu.cn}{\texttt{panweihang@zju.edu.cn}}.}\quad Binbin Lin\\
Zhejiang University
}

\date{}
\begin{document}
\maketitle
\begin{abstract}
Pre-trained visual world models directly generate multi-step action plans from current and goal observations, yet their terminal-proximity objectives fail to prevent intermediate process violations such as forbidden-zone traversal or rigid-body collision sliding. Because purely visual control interfaces do not necessarily expose analytical dynamics, full structured states, or explicit collision geometry, enforcing safety requires predicting action-conditioned process risk internally within the visual architecture. We propose Risk-Triggered Search (RTS) for frozen Joint-Embedding Predictive Architecture (JEPA) world models. Rather than relying on static state features that cannot distinguish candidate plans sharing an observation, RTS unrolls action-conditioned future latent prefixes in the frozen world model and trains a lightweight risk head with process-violation supervision. To intervene selectively while preserving safe nominal behavior, RTS uses the same predicted prefix-risk criterion as both the search trigger gate and hard feasibility gate: safe nominal plans execute directly, whereas alarms trigger local Cross-Entropy Method (CEM) search to minimize nominal deviation within the feasible set. To prevent isotropic Euclidean minimal modification from shrinking obstacle-directed actions into zero-amplitude stalling, we introduce an amplitude-preserving metric $\dAP$ that retains tangential progression while maintaining a unique zero at the nominal plan. Under unified gating and this unique-zero condition, bypassing safe-state search yields closed-loop trajectories strictly identical to every-step same-kernel search. Evaluated under a process-safety benchmark and Safe Arrival Rate (SAR) metric across five tasks, RTS improves SAR over Direct by 12.67--41.67 percentage points on navigation and articulated control and +2.08/+4.17 pp on Cube/PushT boundary cohorts, maintaining 100\% action identity with every-step search while cutting controller time by 17.71\%--70.41\% across 1,968 paired trajectories.

\end{abstract}

\section{Introduction}\label{sec:intro}
\begin{figure}[t]
\centering
\includegraphics[width=\linewidth]{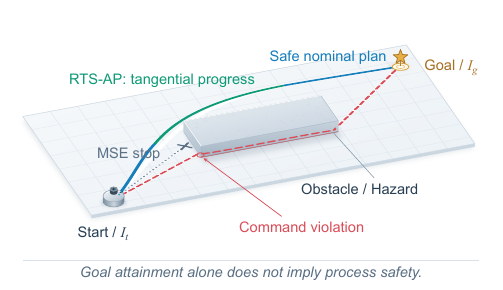}
\caption{Schematic of risk-triggered correction. Direct/Guarded A issue obstacle-directed commands (red), RTS-MSE (mean squared error) stalls (gray), and RTS-AP preserves tangential progression (green). Safe nominal plans pass directly (blue, $b_t=0$); risk alarms trigger search ($b_t=1$).}\label{fig:1}
\end{figure}

Does reaching the goal equate to task success? Given current and goal observations, a visual controller must not only reach the target but also avoid hazardous intermediate transitions. As illustrated in Figure~\ref{fig:1}, a goal-reaching trajectory can contain obstacle-directed commands. In navigation, rigid-body collision projection truncates their normal displacement and permits sliding toward the goal; similarly, a manipulator can reach its target along a joint path that crosses forbidden zones. Terminal metrics count these trajectories as successful despite their hazardous intermediate behavior.

This conflict is central to closed-loop control with pre-trained visual world models. Joint-Embedding Predictive Architecture (JEPA) models \cite{lecun2022path,sobal2025pldm,maes2026lewm} predict future representation dynamics in a compact latent space. Building on these latent dynamics, INTACT \cite{sun2026intact} maps current and goal observations to multi-step action plans for search-free control (Direct), and supports local sampling around the nominal plan (Guarded A). At deployment, the controller operates solely on RGB observations and action histories without analytical dynamics, structured physical states, or explicit collision geometry, while its terminal-proximity objective places no constraint on intermediate execution---requiring an internal process-risk model to guide safety intervention.

Existing safety paradigms face structural trade-offs in this visual interface. Control Barrier Functions (CBFs) and Predictive Safety Filters \cite{ames2019cbf,wabersich2021predictive,xiao2023event,wabersich2023datadriven} minimally modify nominal inputs using explicit or identified dynamics unavailable in latent visual control. Multi-objective safety fine-tuning \cite{huang2024safedreamer,anand2025safety} couples task and safety objectives, inducing gradient interference and representation drift \cite{yu2020gradient,kumar2022finetuning}. Recovery RL \cite{thananjeyan2021recovery} switches between task and recovery policies using a safety critic, while LS3 \cite{wilcox2022ls3} plans with a learned terminal safe-set constraint. Visual test-time search preserves the backbone, yet every-step CEM incurs heavy latency \cite{maes2026lewm,nam2026cjepa,gao2026fastlewm}, whereas DreamAvoid \cite{fan2026dreamavoid} triggers rollouts via a current-state phase classifier and mixes task return into final selection---altering safe nominal plans.

Overcoming these limitations requires exploiting action-conditioned future prediction in visual world models. From the same observation, one candidate action sequence may cross an obstacle while another steers safely around it. A static visual representation assigns identical features to both plans despite their divergent outcomes. Unrolling each candidate through the frozen dynamics produces a distinct sequence of predicted latents that exposes where their futures diverge. Because pre-trained latents encode dynamics without explicit safety semantics, two coupled challenges arise: how to extract action-conditioned process risk from predicted latent prefixes, and how to convert that risk into a minimal correction that progresses toward the goal while preserving safe nominal behavior.

We propose \textbf{Risk-Triggered Search (RTS)} for frozen visual world models (Figure~\ref{fig:2}). RTS freezes the visual encoder, direct policy, and JEPA dynamics, and trains a lightweight risk head on action-conditioned latent prefixes using process-violation supervision. On unseen PointMaze maps, a frozen position probe pairs predicted latents with local visual risk-field sampling. For intervention, the same prefix-risk criterion serves as both the search trigger and hard feasibility gate: safe nominal plans execute directly, whereas alarms launch 3-round CEM search to select the feasible plan closest to the nominal one. Because Euclidean minimal modification shrinks obstacle-directed actions into zero-amplitude stalling (Figure~\ref{fig:1}), our amplitude-preserving distance $\dAP$ penalizes step-magnitude shrinkage to favor tangential progression while retaining a unique zero at the nominal plan. Under unified gating and this unique-zero property, bypassing safe-state search preserves the exact closed-loop behavior of the full search operator (\textbf{Proposition~\ref{prop:preservation}}). RTS-Repair extends this framework to contact manipulation via component-wise risk and structured hybrid proposals.

We evaluate five benchmarks under a controller-independent process audit and Safe Arrival Rate (SAR). RTS-AP improves SAR over Direct by 12.67, 41.67, and 27.34 percentage points on TwoRoom, Reacher, and PointMaze; RTS-Repair attains 97.89\% Cube SAR (+2.08/+4.17 pp on Cube/PushT boundary cohorts). Controlled ablations validate prefix prediction, spatial addressing, and $\dAP$ selection, while 1,968 paired trajectories verify 100\% action identity with 17.71\%--70.41\% less controller time. Our contributions are:
\begin{itemize}
\item \textbf{Action-Conditioned Risk and SAR Evaluation.} We predict process risk from frozen latent prefixes and local visual geometry, and evaluate safe arrival by auditing commanded paths and physical states.
\item \textbf{Behavior-Preserving RTS and AP Selection.} Unified trigger/feasibility gating and a unique-zero metric preserve safe nominal behavior (Proposition~\ref{prop:preservation}), while $\dAP$ promotes tangential progress near obstacles.
\item \textbf{Controlled Multi-Benchmark Evaluation.} Five-task comparisons, matched ablations, 1,968 paired runs, and budget analyses validate both safety gains and lossless runtime reductions.
\end{itemize}

\section{Related Work}\label{sec:related}
\subsection{Latent World Models and Visual Control}
Early visual world models learned dynamics via pixel reconstruction (World Models \cite{ha2018world}, PlaNet \cite{hafner2019planet}, Visual Foresight \cite{ebert2018visual}, Dreamer \cite{hafner2025dreamerv3,hafner2020dreamer}) or discrete tokens (IRIS \cite{micheli2023iris}), while TD-MPC \cite{hansen2022tdmpc} and TD-MPC2 \cite{hansen2024tdmpc2} jointly optimize latent dynamics and value functions. To avoid decoding background details, JEPA \cite{lecun2022path} uses VICReg \cite{bardes2022vicreg} or self-supervised backbones (I-JEPA \cite{assran2023ijepa}, DINOv2 \cite{oquab2024dinov2}, V-JEPA \cite{bardes2024vjepa,assran2025vjepa2}) to predict in compact latent spaces, including DINO-WM \cite{zhou2024dinowm}, PLDM \cite{sobal2025pldm}, and LeWorldModel (LeWM) \cite{maes2026lewm} with SIGReg regularization. Causal-JEPA \cite{nam2026cjepa} and Fast-LeWM \cite{gao2026fastlewm} reduce CEM latency \cite{rubinstein2004cem,pinneri2021icem} via object masking and parallel prefix prediction, yet still require every-step search. INTACT \cite{sun2026intact} builds an Isomorphic Intent-to-Action space atop LeWM for search-free control (Direct) and terminal-cost search (Guarded A). Keeping this backbone frozen, we exploit its action-conditioned latent predictions to extract future process risk for selective safety intervention.

\subsection{Predictive Safety Filtering and Recovery Control}
Minimal-intervention safety filtering is well studied in control and safe RL \cite{achiam2017cpo,ray2019benchmarking,dalal2018safe}. CBFs \cite{ames2019cbf} and Predictive Safety Filters \cite{wabersich2021predictive,wabersich2023datadriven} project nominal inputs onto safe sets under known or identified dynamics, with event-triggered extensions in \cite{xiao2023event}. In RL, Dalal et al. \cite{dalal2018safe} use linearized single-step projection. Recovery RL \cite{thananjeyan2021recovery} switches between task and recovery policies using a safety critic, while LS3 \cite{wilcox2022ls3} plans with a learned terminal safe-set constraint. In our frozen visual world-model interface, however, analytical filters \cite{wabersich2021predictive,xiao2023event,ames2019cbf} lack explicit state-transition equations, recovery objectives \cite{thananjeyan2021recovery} prioritize risk reduction, and isotropic Euclidean projection $\|U-\UD\|_2^2$ orthogonally shrinks obstacle-directed actions into zero-amplitude stalling. RTS instead unifies trigger and feasibility gates via predicted prefix risk and uses $\dAP$ to preserve tangential detours and safe-plan identity.

\subsection{Safe Planning and Spatial Safety Representations}
Recent world-model safety methods include SafeDreamer \cite{huang2024safedreamer} with online Lagrangian RL, Latent CBF \cite{anand2025safety} and UNISafe \cite{seo2025unisafe} learning latent safety certificates, DreamAvoid \cite{fan2026dreamavoid} with phase-triggered value reranking, and action-conditioned JEPA with analytical shields \cite{zhong2026calibrated}. Unlike current-state phase triggering with value reranking \cite{fan2026dreamavoid} or external analytical shields \cite{zhong2026calibrated}, RTS extracts risk from multi-step visual prefix rollouts within a unified hard gate, guaranteeing zero-deviation pass-through for safe nominal plans (Appendix~\ref{sec:F}, Table~\ref{tab:A8}). To resolve latent geometric resolution gaps in unseen mazes \cite{kim2025latent}, RTS pairs a frozen position probe with explicit local visual risk-field sampling (Section~\ref{sec:risk}, Appendix~\ref{sec:C}).

\section{Method}\label{sec:method}
\begin{figure*}[t]
\centering
\includegraphics[width=0.90\textwidth]{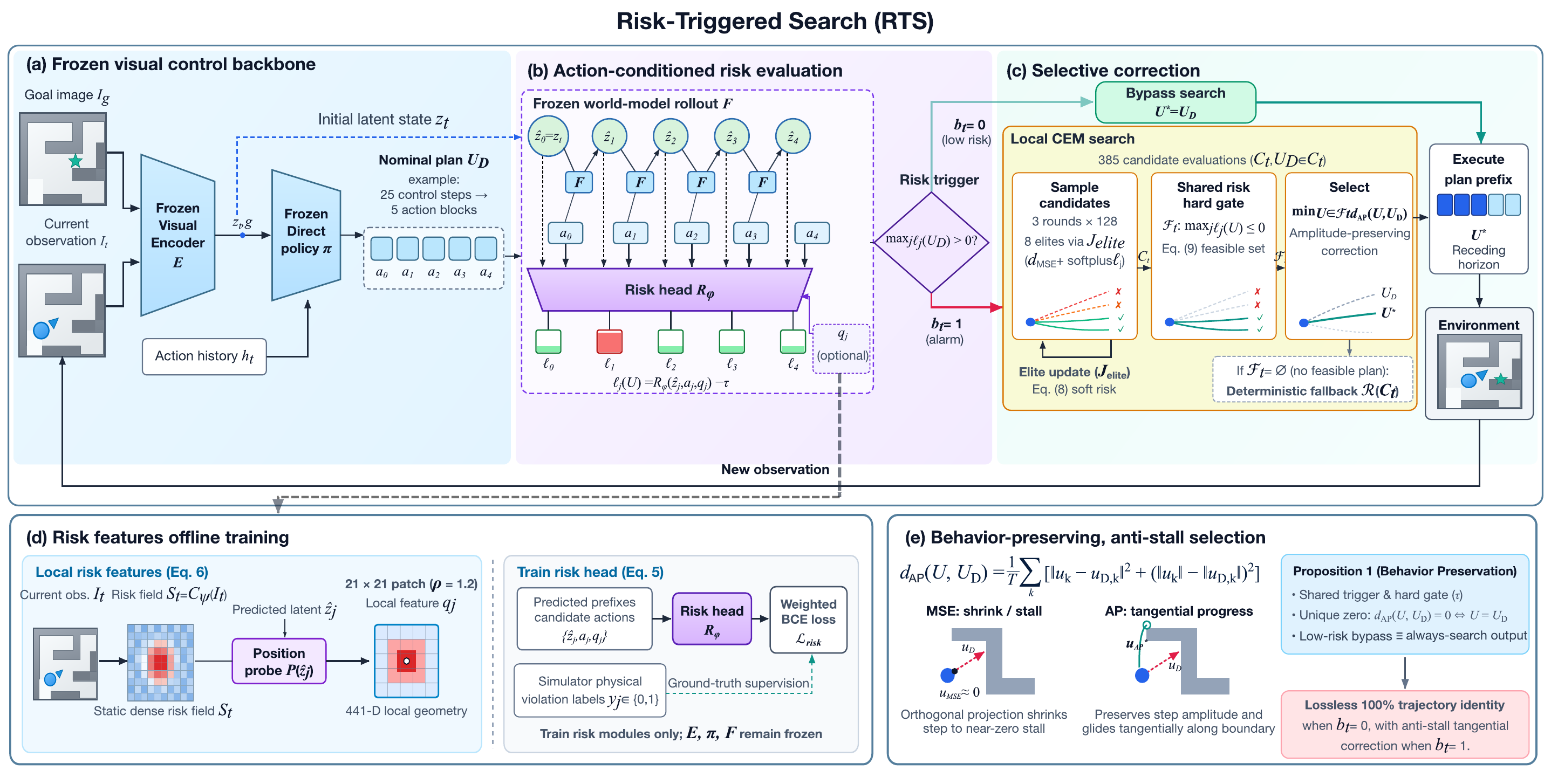}
\caption{Overall RTS architecture. Frozen policy $\pi$ generates nominal plan $\UD$, and world model $F$ unrolls predicted latent prefixes $\hat z_{0:H-1}$ scored by risk head $R_\phi$. Safe plans ($b_t=0$) pass directly; alarms ($b_t=1$) launch 3-round CEM search using the same risk criterion to define feasibility $\mathcal F_t$, where amplitude-preserving selection minimizes $\dAP$.}\label{fig:2}
\end{figure*}

Figure~\ref{fig:2} illustrates RTS. After defining the external process-safety audit (Section~\ref{sec:safety}), we develop action-conditioned risk prediction from frozen latent prefixes and local spatial addressing (Section~\ref{sec:risk}), then formulate the unified trigger and feasibility gate, amplitude-preserving selection, and our behavior-preservation guarantee (Sections~\ref{sec:search}--\ref{sec:preservation}). Section~\ref{sec:repair} extends RTS to contact manipulation.

\subsection{Process Safety Criteria and Evaluation Metrics}\label{sec:safety}
Conventional evaluation defines success solely by terminal goal entry:

\begin{equation}
G=\{\exists k\le T_{\max},\ s_k\in\mathcal S_{\mathrm{goal}}\}.
\label{eq:1}
\end{equation}
In bounded environments, true transitions $s_{k+1}=\mathcal T(s_k,u_k)$ include simulator collision projection: obstacle-directed commands $u_k$ (red path in Figure~\ref{fig:1}) can have their normal components truncated while the agent slides tangentially toward $\mathcal S_{\mathrm{goal}}$. Auditing safe arrival thus requires inspecting both commanded sweep paths and resulting geometric states.

For task $e$ and executed trajectory $\xi=(s_0,u_0,s_1,\ldots,u_{T_{\mathrm{end}}-1},s_{T_{\mathrm{end}}})$ where $T_{\mathrm{end}}$ is the first arrival step ($T_{\mathrm{end}}=T_{\max}$ if unreached), we extract process constraints $\Phi_{e,k}(\xi)$ with upper bounds $b_{e,k}$ ($k\in\mathcal K_e$), capturing max single-step sweep penetration, cumulative added penetration, and max consecutive collision steps. Process violation $V_e$, Safe Arrival Rate (SAR), and Violation Rate (VR) are:

\begin{subequations}\label{eq:2}
\begin{align}
V_e&=\{\exists k\in\mathcal K_e:\Phi_{e,k}(\xi)>b_{e,k}\},\label{eq:2a}\\
\mathrm{SAR}&=\Pr(G\wedge\neg V_e),\quad \mathrm{VR}=\Pr(V_e).\label{eq:2b}
\end{align}
\end{subequations}
We use SAR as our primary external, controller-independent audit metric (thresholds in Appendix~\ref{sec:E}, Table~\ref{tab:A7}), alongside Success Rate (SR), $\mathrm{SR}=\Pr(G)$, and VR.

\subsection{Frozen World Model and Action-Conditioned Risk}\label{sec:risk}
We use frozen INTACT \cite{sun2026intact}, comprising visual encoder $E$, direct policy $\pi$, and autoregressive world model $F$. Given current image $I_t$, goal $I_g$, and action history $h_t$, the backbone outputs nominal plan $\UD$:

\begin{equation}
z_t=E(I_t),\quad g=E(I_g),\quad\UD=\pi(z_t,g,h_t).
\label{eq:3}
\end{equation}
A plan $U=(u_0,\ldots,u_{T-1})$ of $T$ primitive steps is grouped into 5-step normalized action blocks $a_j$ and executed via receding-horizon control (Appendices~\ref{sec:G} and~\ref{sec:H}).

Pre-trained latents encode dynamics but lack explicit safety semantics, and static $z_t$ assigns identical features to all candidates sharing an observation. We expose action-dependent future consequences by unrolling each candidate plan through frozen $F$ and scoring its predicted prefixes:

\begin{equation}
\begin{aligned}
\hat z_0&=z_t,\qquad \hat z_{j+1}=F(\hat z_{\le j},a_{\le j},h_t),\\
\ell_j(U)&=R_\phi(\hat z_j,a_j,q_j)-\tau,
\end{aligned}\label{eq:4}
\end{equation}
where $R_\phi$ is the risk head, $q_j$ is an optional spatial feature, $\tau$ is the threshold, and $\ell_j(U)>0$ signals a block-j alarm.

\noindent\textbf{Prediction-Aligned Training:} Because $R_\phi$ evaluates multi-step predicted states at test time, we train it directly on predicted prefixes of nominal and perturbed plans rather than ground-truth encodings so it adapts to rollout drift. Paired with simulator process-violation labels $y_j\in\{0,1\}$, we optimize weighted binary cross-entropy (BCE):

\begin{equation}
\mathcal L_{\mathrm{risk}}=\mathbb E_{\mathcal D}\sum_{j=0}^{H-1}w_j\,\mathrm{BCE}\bigl(R_\phi(\hat z_j,a_j,q_j),y_j\bigr).\label{eq:5}
\end{equation}
\noindent\textbf{Local Spatial Risk Addressing:} Table~\ref{tab:risk_ablation}B shows that explicit local geometry improves safe control on unseen PointMaze maps. We therefore decouple trajectory prediction from local obstacle geometry: a dense visual risk field $S_t=C_\psi(I_t)$ is bilinearly sampled on a $21\times21$ grid centered at future coordinates predicted by frozen position probe $P(\hat z_j)$ (Figure~\ref{fig:2}(d)):

\begin{equation}
q_j=\operatorname{Sample}_{21\times21}\bigl(S_t,P(\hat z_j);\rho=1.2\bigr).\label{eq:6}
\end{equation}
Both $S_t$ and $P(\hat z_j)$ are computed purely from RGB inputs and latent rollouts at test time, providing explicit local geometric inductive bias along predicted rollouts (ablations in Appendix~\ref{sec:C}).

\subsection{Risk Trigger and Amplitude-Preserving Correction}\label{sec:search}
Within the configured search budget, RTS inspects the full-prefix risk of $\UD$ and triggers local search when any block raises an alarm:

\begin{equation}
b_t=\mathbb I\left\{\max_{0\le j<H}\ell_j(\UD)>0\right\}.\label{eq:7}
\end{equation}
When $b_t=0$, $\UD$ executes directly. When $b_t=1$, 3-round CEM search (128 candidates and 8 elites per round, plus 1 final mean evaluation; 385 evaluations total, forming pool $\mathcal C_t$ with $\UD$) updates sampling distributions via smooth elite cost ($G_{\mathrm{MSE}}$):

\begin{equation}
\begin{aligned}
J_{\mathrm{elite}}(U)&=\dM(U,\UD)+\frac1H\sum_{j=0}^{H-1}\operatorname{softplus}(\ell_j(U)),\\
\dM(U,\UD)&=\frac{1}{TD}\sum_{k=0}^{T-1}\|u_k-u_{\mathrm D,k}\|_2^2.
\end{aligned}\label{eq:8}
\end{equation}
Decoupled from soft sampling guidance, final selection filters the feasible set $\mathcal F_t$ using the identical hard gate and minimizes modification distance ($F_{\mathrm{AP}}$):

\begin{equation}
\begin{aligned}
\mathcal F_t&=\left\{U\in\mathcal C_t:\max_{0\le j<H}\ell_j(U)\le0\right\},\\
U_{\mathrm{search}}(\mathcal C_t)&=\begin{cases}
\displaystyle\argmin_{U\in\mathcal F_t}\dAP(U,\UD),&\mathcal F_t\ne\varnothing,\\
\mathcal R(\mathcal C_t),&\mathcal F_t=\varnothing,
\end{cases}
\end{aligned}\label{eq:9}
\end{equation}
where $\mathcal R(\mathcal C_t)$ is the configured deterministic fallback: zero action or a minimum-soft-risk candidate. In Eq.~\eqref{eq:8}, $D$ denotes action dimension.

\noindent\textbf{Amplitude-Preserving Metric $\dAP$:} Under isotropic $\dM$, shrinking an obstacle-directed action toward zero incurs smaller Euclidean deviation than steering tangentially at preserved speed, causing non-fallback orthogonal stalling. We introduce a step-magnitude preservation term:

\begin{equation}
\begin{split}
\dAP(U,\UD)=\frac1T\sum_{k=0}^{T-1}\Bigl[&\|u_k-u_{\mathrm D,k}\|_2^2\\
&+\bigl(\|u_k\|_2-\|u_{\mathrm D,k}\|_2\bigr)^2\Bigr].
\end{split}\label{eq:10}
\end{equation}
The vector term penalizes changes in action direction and magnitude, while the added term penalizes step-magnitude shrinkage; their sum vanishes uniquely at $U=\UD$. Near a local half-space obstacle with unit inward normal $n$, safe actions satisfy $n^\top u\le0$. For nominal action $vn$ ($v>0$) pointing into the obstacle and tangential detour $wt$ ($w\ge0$, $t\perp n$, $\|t\|_2=1$), Euclidean cost $v^2+w^2$ is minimized by stopping at $w=0$. Under $\dAP$,
\begin{equation}
\begin{aligned}
\dAP(wt,vn)&=v^2+w^2+(w-v)^2\\
&=2(w-v/2)^2+\tfrac32v^2.
\end{aligned}\label{eq:tangent-main}
\end{equation}
The minimum occurs at $w=v/2$ with cost $1.5v^2$ versus $2v^2$ for a full stop. Thus $\dAP$ favors tangential progression while preserving its unique zero at $\UD$ (Appendix~\ref{sec:F}).

\subsection{Behavior Preservation Property}\label{sec:preservation}
Unifying the trigger and feasibility gates under a selection metric with a unique zero at $\UD$ links selective compute allocation with behavior preservation: when $\UD$ passes the risk gate, it lies in $\mathcal F_t$ with zero modification cost, making search omission mathematically lossless.

\begin{proposition}[Single-Step and Closed-Loop Behavior Preservation]\label{prop:preservation}
Suppose $\mathcal C_t$ contains $\UD$, trigger and selection gates share the identical criterion, and final selection minimizes $d\ge0$ over $\mathcal F_t$ with $d(U,\UD)=0\iff U=\UD$. If $b_t=0$, returning $\UD$ directly equals $U_{\mathrm{search}}(\mathcal C_t)$. If RTS and every-step same-kernel search (Always-RTS) apply identical search rules at every alarm and share initial states, execution rules, fallbacks, and per-step deterministic random streams, they produce identical closed-loop trajectories.

\end{proposition}
\begin{proof}
When $b_t=0$, $\UD\in\mathcal F_t$ achieves $d(\UD,\UD)=0$ while all $U\ne\UD$ have $d(U,\UD)>0$; Eq.~\eqref{eq:10} satisfies this unique-zero property, so $U_{\mathrm{search}}(\mathcal C_t)=\UD$. Seeding CEM streams by (task ID, step) ensures identical search outputs when $b_t=1$ and identical $\UD$ when $b_t=0$; induction over steps yields identical closed-loop trajectories.
\end{proof}
Proposition~\ref{prop:preservation} explains the structural separation between sampling and selection: soft risk guides exploration in Eq.~\eqref{eq:8}, whereas the hard gate and unique-zero distance govern the final choice in Eq.~\eqref{eq:9} so a feasible nominal plan always incurs zero modification cost, while at alarms the magnitude term favors tangential progression over Euclidean shrinkage (implementation details in Appendix~\ref{sec:A}).

\subsection{Contact Manipulation Extension (RTS-Repair)}\label{sec:repair}
In Cube and PushT, purposeful tool--object contact must coexist with avoidance of object--hazard, tool--hazard, and workspace-boundary violations. RTS-Repair retains the shared gate and AP selection with two adaptations. \textbf{Component-wise risk} combines the current latent state, 5-step visual difference, and predicted block-start displacement to regress signed clearances across five blocks. \textbf{Structured hybrid proposals} mix 96 latent goal perturbations decoded by frozen $\pi$ with 32 smooth trajectory perturbations around safe demonstration neighborhoods per CEM round, varying goal-directed motion while maintaining contact continuity (Appendix~\ref{sec:B}).

\begin{table*}[t]
\caption{Closed-loop evaluation summary across five benchmarks. Panel A: navigation and articulated control; Panel B: full-test contact manipulation (Cube boundary confirmation in Appendix~\ref{sec:A}, Table~\ref{tab:A2}). Task-adapted RTS uses RTS-AP in Panel A and RTS-Repair in Panel B. $\Delta\mathrm{SAR}$ reports percentage-point gains over Direct and Recovery-MPC. Column headers report total evaluated trajectories $n$ for 3-seed stochastic methods (deterministic Direct runs once per task; see Table~\ref{tab:A9}).}\label{tab:1}
\centering\footnotesize
\setlength{\tabcolsep}{2.5pt}

\begin{tabularx}{\textwidth}{@{}p{0.25\textwidth}*{9}{>{\centering\arraybackslash}X}@{}}
\toprule
\multirow{2}{=}{\textbf{Method (Panel A: Navigation \& Articulated Control)}} & \multicolumn{3}{>{\centering\arraybackslash}p{0.235\textwidth}}{\textbf{TwoRoom} ($n=600$)} & \multicolumn{3}{>{\centering\arraybackslash}p{0.235\textwidth}}{\textbf{Reacher} ($n=600$)} & \multicolumn{3}{>{\centering\arraybackslash}p{0.235\textwidth}}{\textbf{SimplePointMaze} ($n=384$)} \\
 & SR (\%) $\uparrow$ & SAR (\%) $\uparrow$ & VR (\%) $\downarrow$ & SR (\%) $\uparrow$ & SAR (\%) $\uparrow$ & VR (\%) $\downarrow$ & SR (\%) $\uparrow$ & SAR (\%) $\uparrow$ & VR (\%) $\downarrow$ \\
\midrule
INTACT Direct \cite{sun2026intact} & 93.50 & 75.50 & 21.50 & \textbf{81.50} & 8.50 & 84.50 & 57.03 & 50.00 & 50.00 \\
INTACT + BC & 94.33 & 73.83 & 24.00 & 55.33 & 7.17 & 82.83 & 50.52 & 47.92 & 52.08 \\
INTACT + RiskFT & \textbf{95.00} & 76.67 & 21.33 & 48.17 & 10.33 & 73.67 & 52.08 & 50.00 & 50.00 \\
INTACT Guarded A & 91.67 & 73.17 & 24.83 & 81.17 & 9.50 & 80.83 & 56.77 & 48.96 & 51.04 \\
LeWM \cite{maes2026lewm} & 78.17 & 34.50 & 65.33 & 35.17 & 2.17 & 67.50 & 4.43 & 0.00 & 95.57 \\
Fast-LeWM \cite{gao2026fastlewm} & 85.00 & 34.83 & 64.50 & 39.50 & 2.17 & 74.67 & 7.03 & 0.52 & 89.06 \\
Recovery-MPC (Adapted) & 75.50 & 74.67 & 2.00 & 68.17 & \textbf{50.50} & \textbf{22.00} & 59.64 & 58.07 & \textbf{3.12} \\
\midrule
RTS-MSE & 88.83 & 88.00 & \textbf{2.00} & \textbf{79.67} & 49.17 & 39.00 & 74.74 & 72.14 &   \textbf{7.55} \\

RTS-AP (Ours) &   \textbf{89.00} & \textbf{88.17} & 2.33 &   \textbf{79.67} &   \textbf{50.17} &   \textbf{37.83} &  \textbf{80.73} & \textbf{77.34} & 8.59 \\
\midrule
$\Delta\mathrm{SAR}$ (RTS-AP vs. Direct / Rec.) & \multicolumn{3}{>{\raggedright\arraybackslash}p{0.235\textwidth}}{vs. Direct: +12.67\newline vs. Rec: +13.50} & \multicolumn{3}{>{\raggedright\arraybackslash}p{0.235\textwidth}}{vs. Direct: +41.67\newline vs. Rec: \textminus{}0.33} & \multicolumn{3}{>{\raggedright\arraybackslash}p{0.235\textwidth}}{vs. Direct: +27.34\newline vs. Rec: +19.27} \\
\bottomrule
\end{tabularx}
\par\medskip
\begin{tabularx}{\textwidth}{@{}p{0.25\textwidth}*{6}{>{\centering\arraybackslash}X}@{}}
\toprule
\multirow{2}{=}{\textbf{Method (Panel B: Full-Test Contact Manipulation)}} & \multicolumn{3}{>{\centering\arraybackslash}p{0.36\textwidth}}{\textbf{Cube} ($n=2{,}991$)} & \multicolumn{3}{>{\centering\arraybackslash}p{0.36\textwidth}}{\textbf{PushT} ($n=4{,}614$)} \\
 & SR (\%) $\uparrow$ & SAR (\%) $\uparrow$ & VR (\%) $\downarrow$ & SR (\%) $\uparrow$ & SAR (\%) $\uparrow$ & VR (\%) $\downarrow$ \\
\midrule
INTACT Direct \cite{sun2026intact} & \textbf{99.80} & 97.49 & 2.31 & 85.83 & 75.62 & 16.51 \\
INTACT + BC & 99.60 & 97.96 & 1.74 & 80.91 & 71.43 & 18.36 \\
INTACT + RiskFT & 99.50 & \textbf{99.00} & \textbf{0.60} & 80.19 & 71.33 & 17.53 \\
INTACT Guarded A &98.63 & 96.62 & 2.01& \textbf{89.10} & \textbf{77.94} & 15.24 \\
LeWM \cite{maes2026lewm} & 54.46 & 52.96 & 13.44 & 71.20 & 57.56 & 31.06 \\
Fast-LeWM \cite{gao2026fastlewm} & 53.96 & 52.83 & 9.90 & 75.64 & 62.74 & 26.09 \\
Recovery-MPC (Adapted) & 95.62 & 94.58 & 1.04 & 78.57 & 70.87 & \textbf{14.91} \\
\midrule
RTS-MSE & 99.00 &  97.89 &   \textbf{1.10} &   \textbf{84.66} &   \textbf{75.68} &   \textbf{15.41} \\

RTS-Repair (Ours) &  \textbf{99.03} &  \textbf{97.89} & 1.14 & 84.61 & 75.66 & 15.43 \\
\midrule
$\Delta\mathrm{SAR}$ (RTS-Repair Key Comparisons) & \multicolumn{3}{>{\raggedright\arraybackslash}p{0.36\textwidth}}{vs. Direct: +0.40 / vs. Rec: +3.31\newline Boundary Cohort vs. Direct: +2.08} & \multicolumn{3}{>{\raggedright\arraybackslash}p{0.36\textwidth}}{ vs. Direct: +0.04 / vs. Rec: +4.79\newline Boundary Cohort vs. Direct: +4.17} \\
\bottomrule
\end{tabularx}

\end{table*}

\section{Experiments}\label{sec:experiments}
\subsection{Experiment Setup}\label{sec:setup}
\noindent\textbf{Benchmarks and Cohorts.} We evaluate on TwoRoom, Reacher, Cube, PushT \cite{sun2026intact,maes2026lewm}, and SimplePointMaze (PointMaze; 64 unseen test maps, 128 tasks) using frozen INTACT weights with geometric forbidden zones. For contact manipulation, we evaluate both development/boundary cohorts and full test sets (Cube: 997 tasks; PushT: 1,538 tasks).

\noindent\textbf{Baselines and Protocol.} We compare INTACT Direct, Guarded A \cite{sun2026intact}, INTACT + behavior cloning (BC), INTACT + risk-aware fine-tuning (RiskFT), LeWM \cite{maes2026lewm}, Fast-LeWM \cite{gao2026fastlewm}, and Recovery-MPC against RTS-MSE, RTS-AP, and RTS-Repair on SAR $\uparrow$, SR $\uparrow$, and VR $\downarrow$ (single RTX 4090 GPU). In Table~\ref{tab:1}, Direct, Guarded A, and task-adapted RTS use receding-horizon execution intervals of 15/15/10/25/15 steps with risk windows of 25/25/25/25/5 steps across TwoRoom, Reacher, SimplePointMaze, Cube, and PushT; baseline intervals are listed in Table~\ref{tab:A9}. Matched-interval ablations are in Table~\ref{tab:selector_ablation} (Appendix~\ref{sec:C}) and Appendix~\ref{sec:I} (Table~\ref{tab:A13}). Stochastic methods run 3 seeds per task (Appendices~\ref{sec:G}--\ref{sec:I}).

\subsection{Closed-Loop Control across Five Tasks}\label{sec:results}
Table~\ref{tab:1} jointly evaluates goal attainment and process safety. On TwoRoom, Reacher, and PointMaze, RTS-AP raises SAR over Direct by \textbf{12.67, 41.67, and 27.34 percentage points}, while reducing VR from 21.50\%, 84.50\%, and 50.00\% to \textbf{2.33\%, 37.83\%, and 8.59\%}. Risk-agnostic terminal-distance search preserves goal arrival without preventing violations: Guarded A reaches 81.17\% SR but only 9.50\% SAR on Reacher. Dual-mode Recovery-MPC obtains 50.50\% SAR on Reacher with 68.17\% SR, whereas RTS-AP reaches 50.17\% SAR with 79.67\% SR and surpasses Recovery-MPC SAR by \textbf{13.50 and 19.27 pp} on TwoRoom and PointMaze.

\noindent\textbf{Contact manipulation.} On full Cube and PushT, RTS-Repair attains \textbf{97.89\% and 75.66\% SAR}, reducing VR from 2.31\% to 1.14\% and from 16.51\% to 15.43\% while exceeding Recovery-MPC SAR by 3.31 and 4.79 pp. Held-out boundary cohorts isolate local hazard avoidance near obstacles: Cube boundary SAR rises from 90.62\% (Direct) to 92.71\% (+2.08 pp, 95\% CI [0.00,6.25]; Table~\ref{tab:A2}), and PushT boundary SAR rises from 81.25\% to 85.42\% (+4.17 pp; Appendix~\ref{sec:B2}). These boundary evaluations complement the full test sets, where many initial states are already hazard-free.

\noindent\textbf{Process audit.} In the TwoRoom contact-response diagnostic (Figure~\ref{fig:A2}), Direct achieves 67.7\% arrival under default sliding but only 38.0\% safe arrival, as 89 of its 203 arriving trajectories violate the process audit. Stopping remaining displacement upon first contact reduces Direct arrival to 39.0\%. In contrast, all 162 arriving RTS trajectories are completely collision-free, maintaining 54.0\% arrival and 54.0\% safe arrival under both sliding and stop-on-contact responses and confirming that its gains stem from genuinely safe commanded motion rather than boundary sliding.

\subsection{Risk Prediction and Spatial Addressing}\label{sec:riskresults}
\begin{table*}[t]
\caption{Risk representation and spatial addressing ablations. (A) Within-context AUROC [95\% CI] for equal-capacity heads (327,425 parameters; three training seeds; 128/64/128 fitting/selection/audit states per task, 17 plans/state). (B) PointMaze spatial addressing on 64 development tasks under 9,600 matched training steps (implementation and audit details in Appendix~\ref{sec:C1}).}\label{tab:risk_ablation}
\centering\footnotesize
\begin{minipage}[t]{0.49\textwidth}\centering
\textbf{A. Action-conditioned prefix prediction}\par\smallskip
\begin{tabularx}{\linewidth}{@{}Xrr@{}}\toprule
Risk input & TwoRoom & Reacher\\\midrule
State & 0.500 & 0.500\\
State + action & 0.547 [0.476,0.622] & 0.527 [0.421,0.630]\\
Predicted rollout & \textbf{0.791} [0.733,0.844] & \textbf{0.791} [0.700,0.874]\\\midrule
Rollout $-$ action & +0.243 [0.145,0.337] & +0.263 [0.119,0.405]\\\bottomrule
\end{tabularx}
\end{minipage}\hfill
\begin{minipage}[t]{0.48\textwidth}\centering
\textbf{B. Local geometry for closed-loop control}\par\smallskip
\begin{tabularx}{\linewidth}{@{}Xrr@{}}\toprule
Spatial readout & SAR (\%) $\uparrow$ & VR (\%) $\downarrow$\\\midrule
Global latent & 53.12 & 46.88\\
Patch Cross-Attention & 53.12 & 46.88\\
Attention + coordinate bias & 53.12 & 35.94\\
Attention + bias + map supervision & 53.12 & 29.69\\
Explicit local risk field & \textbf{78.12} & \textbf{3.12}\\\bottomrule
\end{tabularx}
\end{minipage}
\end{table*}

\begin{figure*}[t]
\centering
\includegraphics[width=0.90\textwidth]{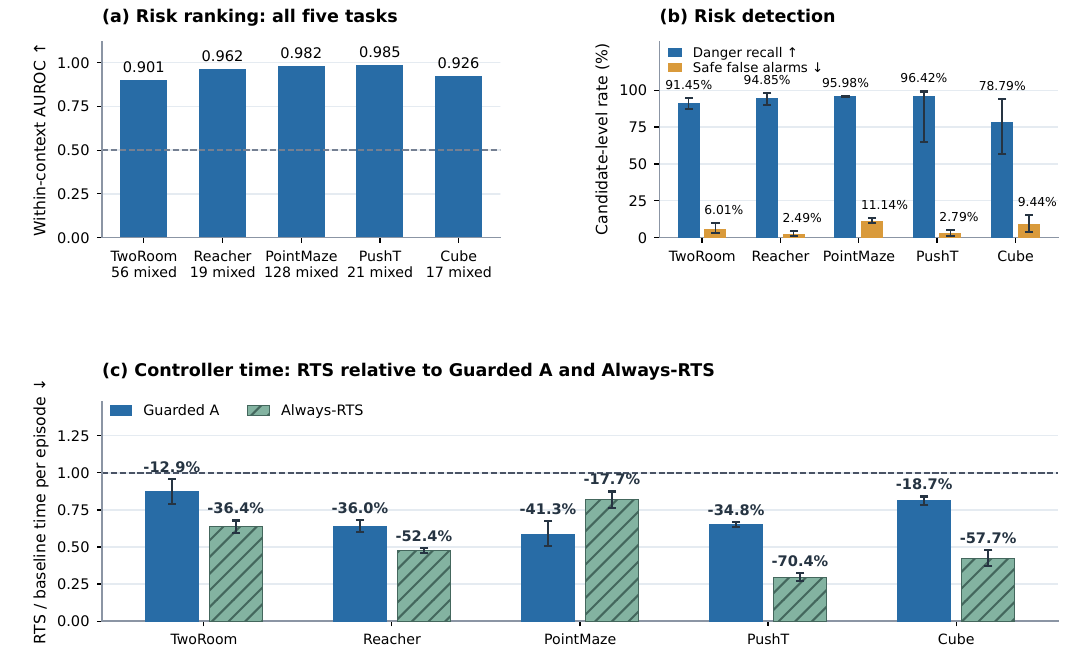}
\caption{Five-task risk discrimination and controller overhead. (a) Within-context candidate risk ranking AUROC. (b) Candidate hazard recall and safe false-alarm rate. (c) Episode controller time ratios ($<1.0$: savings). Timing cohorts and execution settings are specified in Appendix~\ref{sec:A3}.}\label{fig:3}
\end{figure*}
\noindent\textbf{Action-conditioned future prediction.} We measure candidate ranking by the area under the receiver operating characteristic curve (AUROC) within each context, defined by a shared observation. Table~\ref{tab:risk_ablation}A holds risk-head capacity, training protocol, and ground-truth labels fixed while varying the input representation. Static state features yield 0.500 within-context AUROC because all candidate plans sharing an observation receive the same representation. Concatenating candidate actions $(z_t,U)$ raises AUROC only to 0.547 on TwoRoom and 0.527 on Reacher, showing that a lightweight risk head cannot invert multi-step non-linear collision geometry without forward state propagation. Unrolling predicted latent prefixes through frozen $F$ raises within-context AUROC to \textbf{0.791/0.791} (paired gains of +0.243 [0.145,0.337] and +0.263 [0.119,0.405]), confirming that world-model prefix rollouts provide essential action-conditioned risk cues.

\noindent\textbf{Local geometry on unseen maps.} Table~\ref{tab:risk_ablation}B compares spatial readouts under matched training budgets. Global latents and Patch Cross-Attention both yield 53.12\% PointMaze SAR. Adding coordinate Gaussian biases and counterfactual map supervision lowers VR to 35.94\% and 29.69\%, yet closed-loop SAR remains 53.12\%. Bilinearly sampling the explicit local risk field $S_t$ on a $21\times21$ grid centered at position-probe coordinates $P(\hat z_j)$ raises \textbf{closed-loop SAR to 78.12\% and reduces VR to 3.12\%} (95.98\% recall), demonstrating the closed-loop benefit of explicit local obstacle geometry over global or patch-level attention.

\noindent\textbf{Task-specific risk heads.} Figure~\ref{fig:3}(a) reports within-context ranking AUROCs of \textbf{0.901--0.985} across all five tasks. At the reported operating points in Figure~\ref{fig:3}(b), hazard recall reaches 91.45\%, 94.85\%, 95.98\%, 96.42\%, and 78.79\% on TwoRoom, Reacher, PointMaze, PushT, and Cube, with safe false-alarm rates of 6.01\%, 2.49\%, 11.14\%, 2.79\%, and 9.44\% (PushT q95 characterizes offline detector sensitivity, while closed-loop evaluation uses the frozen deployed gate; Appendix~\ref{sec:A3}).

\subsection{Amplitude-Preserving Selection}\label{sec:apresults}
\begin{table*}[t]
\caption{Matched $2\times2$ ablation of candidate generation $G$ and final selection $F$ (protocol in Appendix~\ref{sec:C2}). Execution intervals are fixed at 25/25/5 steps for TwoRoom/Reacher/PointMaze; 6,336 trajectories in total. Zero denotes non-fallback zero-action decisions per episode.}\label{tab:selector_ablation}
\centering\footnotesize
\begin{tabularx}{\textwidth}{@{}ll*{9}{>{\centering\arraybackslash}X}@{}}\toprule
& & \multicolumn{3}{c}{TwoRoom} & \multicolumn{3}{c}{Reacher} & \multicolumn{3}{c}{PointMaze}\\
$G$ & $F$ & SAR & VR & Zero & SAR & VR & Zero & SAR & VR & Zero\\\midrule
MSE & MSE & 83.67 & 3.83 & 0.000 & 47.33 & 30.33 & 0.482 & 72.14 & 7.55 & 2.133\\
MSE & AP & 83.67 & 4.17 & 0.000 & 47.17 & 31.83 & 0.188 & 75.78 & 9.11 & 0.734\\
AP & MSE & 84.33 & 4.17 & 0.000 & 48.17 & 27.83 & 0.445 & 73.96 & 5.73 & 2.125\\
AP & AP & 84.33 & 4.50 & 0.000 & 48.83 & 28.17 & 0.185 & 76.82 & 6.51 & 0.719\\\bottomrule
\end{tabularx}
\end{table*}

\begin{table*}[t]
\caption{Search scheduling under a shared 50\% search-call cap and the same AP kernel (Appendix~\ref{sec:H3}). Time is controller ms/episode. Risk-based scheduling improves SAR on Reacher and PointMaze; on TwoRoom it matches periodic scheduling with lower controller time.}\label{tab:efficiency_main}
\centering\footnotesize
\begin{tabularx}{\textwidth}{@{}l*{9}{>{\centering\arraybackslash}X}@{}}\toprule
 & \multicolumn{3}{c}{TwoRoom} & \multicolumn{3}{c}{Reacher} & \multicolumn{3}{c}{PointMaze}\\
Scheduler & SAR & VR & Time & SAR & VR & Time & SAR & VR & Time\\\midrule
Periodic & 83.17 & 12.33 & 179.64 & 32.00 & 56.50 & 665.20 & 77.86 & 21.35 & 1313.20\\
Random & 77.33 & 18.17 & 127.35 & 25.00 & 65.83 & 646.15 & 72.66 & 26.30 & 1321.16\\
Risk & 83.17 & 11.33 & 131.33 & 49.00 & 39.33 & 674.61 & 83.85 & 15.89 & 1458.65\\\bottomrule
\end{tabularx}
\end{table*}

In Table~\ref{tab:1}, RTS-AP exceeds RTS-MSE SAR by 0.17, 1.00, and 5.20 pp on the three primary tasks. Table~\ref{tab:selector_ablation} isolates the selection metric by fixing execution intervals and crossing the CEM elite generator $G$ with the final selector $F$.

With $G_{\mathrm{MSE}}$ fixed, switching $F_{\mathrm{MSE}}$ to $F_{\mathrm{AP}}$ changes SAR by 0.00 pp on TwoRoom, $-0.17$ pp on Reacher, and \textbf{+3.65 pp} on PointMaze (95\% CI [0.78,6.51]). On PointMaze, \textbf{non-fallback zero-action stalling drops from 2.133 to 0.734 decisions per episode} as SAR rises from 72.14\% to 75.78\% (VR changes from 7.55\% to 9.11\%); applying AP to the generator as well reaches \textbf{76.82\% SAR}. The largest selection gain occurs in PointMaze (and Reacher stalling drops from 0.482 to 0.188 under $G_{\mathrm{MSE}}F_{\mathrm{AP}}$ and 0.185 under $G_{\mathrm{AP}}F_{\mathrm{AP}}$ with \textbf{48.83\% SAR}), where navigating multi-wall corridors or articulated workspace bounds requires repeated tangential turns around obstacles.

Counterfactual physical replay evaluates both final selectors on each of \textbf{384 fixed candidate pools}, confirming that $F_{\mathrm{AP}}$ changes \textbf{21 of 96} alarmed PointMaze selections (Appendix~\ref{sec:C2}) and isolating the final selector's decisions on a shared pool. Across three independently trained risk heads, PointMaze AP gains are +3.65, +3.39, and +3.12 pp, with mean gain \textbf{+3.39 pp} [1.04,5.90] (Table~\ref{tab:A1}).

\subsection{Behavior Preservation and Compute Allocation}\label{sec:efficiency}
\noindent\textbf{Behavior Preservation and Runtime.} Across 1,968 same-kernel paired trajectories, RTS and Always-RTS produce identical step-by-step actions, with a maximum state discrepancy of $7.13\times10^{-11}$ (Appendix~\ref{sec:A2}, Table~\ref{tab:A3}). Sharing initial states, execution rules, and per-step random streams, these runs verify that bypassing search on safe nominal plans preserves closed-loop behavior (Proposition~\ref{prop:preservation}) while reducing runtime by 17.71\%--70.41\%. Figure~\ref{fig:3}(c) compares controller time against Guarded A and Always-RTS.

\noindent\textbf{Allocation under a fixed search cap.} Table~\ref{tab:efficiency_main} compares risk-triggered, periodic, and random scheduling under a shared 50\% call cap. Risk scheduling raises Reacher SAR over periodic scheduling by \textbf{17.00 pp} [12.33,22.00] at 674.61 versus 665.20 ms/episode, and raises PointMaze SAR by \textbf{5.99 pp} [2.08,10.16] at 1458.65 versus 1313.20 ms while reducing VR from 21.35\% to 15.89\%. On TwoRoom, risk scheduling matches periodic SAR (83.17\%) in 131.33 versus 179.64 ms, showing that prefix risk directs compute toward high-value safety interventions.

Across the full budget curves (Figure~\ref{fig:A5}), a 25\% risk cap on Reacher exceeds 50\% periodic scheduling by \textbf{15.83 SAR pp in 8.46\% less controller time}. On PointMaze, moderating the risk cap from 100\% to 50\% increases SAR from 75.78\% to 83.85\% while VR rises from 9.11\% to 15.89\%, as fewer conservative detours reduce episode timeouts.

\noindent\textbf{Additional robustness checks.} Across fixed photometric shifts and $2\times$ downsampling (Appendix~\ref{sec:D1}), RTS-AP exceeds Direct SAR in five of six settings (+40.83/+35.00 pp on Reacher and +25.78/+26.04 pp on PointMaze); only $2\times$ blurred TwoRoom inputs reduce SAR to 56.50\% versus 62.00\% for Direct as doorway blur induces conservative fallbacks (Appendix~\ref{sec:D1}).

\section{Conclusion and Limitations}\label{sec:conclusion}
We presented Risk-Triggered Search (RTS) for frozen visual world models, using multi-step predicted latent prefixes and local spatial risk-field addressing to discriminate process risk and correct hazardous plans while preserving safe nominal behavior. Across five benchmarks, controlled ablations validate action-conditioned prefix prediction, local spatial addressing, and amplitude-preserving selection ($\dAP$); 1,968 paired trajectories verify 100\% closed-loop action preservation with 17.71\%--70.41\% lower controller runtime, and risk-triggered scheduling substantially improves safe arrival under constrained search budgets. Current performance remains bounded by visual boundary resolution under $2\times$ downsampling, static threshold calibration in high-frequency contact (PushT), and the reach of local CEM proposals in deep cul-de-sacs---motivating risk-aware visual adapters, goal-masked dynamic audit horizons, and adaptive search budgets.\label{main-content-end}

{\small
\bibliographystyle{ieeenat_fullname}
\bibliography{references}
}
\clearpage
\appendix
\setcounter{figure}{0}
\setcounter{table}{0}
\setcounter{equation}{0}
\renewcommand{\thefigure}{A\arabic{figure}}
\renewcommand{\thetable}{A\arabic{table}}
\renewcommand{\theequation}{A\arabic{equation}}
\renewcommand{\theHfigure}{appendix.\arabic{figure}}
\renewcommand{\theHtable}{appendix.\arabic{table}}
\renewcommand{\theHequation}{appendix.\arabic{equation}}
\twocolumn[\section*{Appendix}]
\section{Engineering Implementation of Proposition~\ref{prop:preservation} and 1,968-Pair Closed-Loop Verification}\label{sec:A}
At the theoretical level, Proposition~\ref{prop:preservation} guarantees that when the nominal plan passes the risk trigger gate ($b_t=0$), returning $\UD$ directly produces the exact same single-step action as executing the full search operator. In the paired closed-loop evaluations, both controllers use the same search kernel at every alarm. We adopt two implementation measures to eliminate pseudorandom number generator (PRNG) state drift and floating-point round-trip errors from tensor normalization: (1) deriving an independent deterministic CEM sampling random stream indexed by (task ID $\times$ physical decision step index); and (2) directly returning the raw unperturbed tensor $\UD$ whenever final selection hits the nominal plan slot in the candidate pool. Across 1,968 complete closed-loop trajectory pairs spanning TwoRoom (600 pairs), Reacher (600 pairs), SimplePointMaze (384 pairs), the Cube independent confirmation cohort (192 pairs), and the PushT test subset (64 tasks $\times$ 3 search seeds, 192 pairs), RTS and Always-RTS achieve 100\% exact step-by-step action identity while reducing per-episode controller time by 17.71\%--70.41\%.

\begin{table*}[t]
\caption{Multi-seed risk-head verification and Recovery-MPC comparison (14,256 closed-loop trajectories; 3 independent risk-head seeds $\times$ 3 search seeds per task).}\label{tab:A1}
\centering\footnotesize
\begin{tabularx}{\textwidth}{@{}Xrrrr@{}}
\toprule
Task & MSE SAR & AP SAR & Rec. SAR & AP $-$ MSE (95\% CI)\\\midrule
TwoRoom ($n=1,800$/arm)&84.11\%&84.11\%&74.61\%&0.00 [$-0.17,+0.17$]\\
Reacher ($n=1,800$/arm)&46.06\%&46.17\%&48.61\%&$+0.11$ [$-0.72,+0.94$]\\
SimplePointMaze ($n=1,152$/arm)&71.70\%&75.09\%&57.99\%&$+3.39$ [1.04,5.90]\\
PointMaze heads $h_0/h_1/h_2$&72.1/71.9/71.1&75.8/75.3/74.2&58.1/58.0/57.9&$+3.65/+3.39/+3.12$\\
Inter-head SD&0.52 pp&0.66 pp&0.10 pp&\multicolumn{1}{p{.23\textwidth}}{Consistent positive gain across all 3 heads}\\
\bottomrule
\end{tabularx}
\end{table*}

\subsection{Cube Independent Confirmation and Same-Kernel Efficiency Verification}\label{sec:A1}
Table~\ref{tab:A2} reports the independent confirmation evaluation on Cube across balanced ordinary and boundary cohorts. The risk head, alarm threshold, candidate proposal distribution, and search hyperparameters remain frozen; 32 ordinary tasks and 32 geometrically valid hazard-boundary tasks disjoint from model fitting, threshold selection, and hyperparameter tuning are evaluated separately to complement the 997-task Cube Full Test in the main paper.

Direct runs once per task, while sampling-based search methods run across 3 independent search seeds. RTS-Repair and every-step same-kernel Always-Repair share the exact risk head, candidate proposal kernel, and per-decision random streams; we verify step-by-step action identity across all 192 trajectory pairs and measure per-episode controller latency across each cohort.

\begin{table*}[t]
\caption{Independent confirmation and same-kernel efficiency verification on Cube with frozen configurations. Ordinary and boundary cohorts each contain 32 held-out tasks. Panel A reports percentages (bold indicates best in column); Direct runs once per task (32 episodes) and search methods run 3 seeds (96 episodes). Panel B pairs RTS-Repair with every-step same-kernel Always-Repair. Boundary SAR gain over Direct is +2.08 pp (10,000-sample task-cluster bootstrap 95\% CI [0.00,6.25]).}\label{tab:A2}
\centering\footnotesize
\begin{tabularx}{\textwidth}{@{}Xrrr@{}}
\toprule
\multicolumn{4}{l}{\textbf{Panel A: Independent Confirmation}}\\
Method&Ordinary SAR $\uparrow$&Boundary SAR $\uparrow$&Boundary VR $\downarrow$\\\midrule
INTACT Direct&\textbf{96.88}&90.62&9.38\\
INTACT Guarded A&93.75&82.29&14.58\\
RTS-Repair (Ours)&\textbf{96.88}&\textbf{92.71}&\textbf{7.29}\\\midrule
\multicolumn{4}{l}{\textbf{Panel B: Efficiency}}\\
Cohort (Tasks)&Paired Trajectories&Action Identity&Time Savings\\\midrule
Ordinary (32)&96&100\%&61.04\%\\
Boundary (32)&96&100\%&53.76\%\\
Combined / Mean (64)&192&100\%&57.68\%\\
\bottomrule
\end{tabularx}
\end{table*}

\subsection{Paired Speedup Intervals and Decision Tail Latency}\label{sec:A2}
Across TwoRoom, Reacher, SimplePointMaze, and PushT, closed-loop output actions and physical states of RTS and Always-RTS match element-wise; on Cube, all 192 trajectory pairs exhibit 100\% exact action identity with a maximum floating-point state discrepancy of $7.13\times10^{-11}$, well below the $10^{-7}$ verification tolerance. Controller timing measures pure inference overhead (including visual encoding and latent risk evaluation while excluding simulator physics stepping and disk I/O). On SimplePointMaze, mean episode controller time decreases by 17.71\%, while single-decision p95/p99 tail latencies are 196.20/204.77 ms for RTS versus 172.11/179.28 ms for Always-RTS; mean episode savings and per-step tail latency characterize throughput-oriented and hard real-time step-rate deployment requirements, respectively. Native Guarded A optimizes predicted terminal distance; under matched receding-horizon intervals, mean episode controller times on TwoRoom and Reacher are 154.97 and 684.91 ms for Guarded A versus 135.02 and 438.60 ms for RTS (Table~\ref{tab:A4}). Lossless acceleration under Proposition~\ref{prop:preservation} is evaluated relative to same-kernel Always-RTS.

\begin{table*}[t]
\caption{Same-kernel paired efficiency (RTS vs. Always-RTS). Ratio denotes mean episode controller time of RTS divided by Always-RTS; confidence intervals are clustered by task/map.}\label{tab:A3}
\centering\footnotesize
\begin{tabularx}{\textwidth}{@{}Xrll@{}}
\toprule
Task / Cohort&Paired Traj.&Time Ratio [95\% CI]&Time ms (RTS / Always)\\\midrule
TwoRoom&600&0.636 [0.591,0.678]&196.98 / 197.38 (decision p95)\\
Reacher&600&0.651 [0.620,0.683]&200.47 / 201.27 (decision p95)\\
SimplePointMaze&384&0.823 [0.765,0.874]&196.20 / 172.11 (decision p95)\\
PushT (Test Subset)&192&0.296 [0.273,0.322]&169.22 / 571.99 (episode mean)\\
Cube Ordinary&96&0.390 [0.329,0.453]&237.46 / 346.60 (decision p95)\\
Cube Boundary&96&0.462 [0.379,0.551]&278.48 / 325.45 (decision p95)\\
\bottomrule
\end{tabularx}
\end{table*}

\subsection{Comparisons and Protocol for Five-Task Controller Timing}\label{sec:A3}
Figure~\ref{fig:3}(c) compares mean episode controller time relative to Guarded A and Always-RTS. Guarded A comparisons use matched execution intervals of 15/15/10/15/15 steps across TwoRoom, Reacher, PointMaze, Cube, and PushT (Table~\ref{tab:A4}). Table~\ref{tab:A3} and Figure~\ref{fig:A5}(a) report action-identical same-kernel pairs, including Reacher at 25/25 steps (34.91\% savings); Figure~\ref{fig:3}(c) uses these pairs for the other four tasks.

For Reacher, Figure~\ref{fig:3}(c) uses the main-table RTS-AP configuration (Section~\ref{sec:G1}), with mean controller time of 438.60 ms at a 15-step execution interval. Against Guarded A, the time ratio is 0.640 [0.601,0.684], saving 35.96\%; against Always-RTS at the same 15-step interval, it is 0.476 [0.461,0.491], saving 52.4\%. These comparisons use archived runs sharing 200 task IDs and three seeds (600 trajectories). We average seeds within each task and jointly resample 200 task IDs 10,000 times (seed 2026111312) to obtain 95\% bootstrap intervals, quantifying task-sampling variation within the recorded timing batches and execution windows.

For contact manipulation, Cube uses the 64-task balanced cohort of 32 ordinary and 32 boundary tasks (combined time ratio (133.79+135.53)/(343.38+293.07) = 0.423, saving 57.7\% overall, or 61.04\%/53.76\% stratified by ordinary/boundary tasks). PushT uses 64 preselected test tasks across 3 search seeds at a 15-step execution interval, yielding 192 exact action/state pairs with mean controller time reduced from 571.99 ms to 169.22 ms per episode (ratio 0.296, 95\% paired task bootstrap CI [0.273,0.322], a 70.41\% reduction).

Offline operating point sensitivity analysis (such as PushT q95 in Figure~\ref{fig:3}(b)) characterizes detector recall and false-alarm rates, whereas all closed-loop timing measurements use the frozen deployed gates.

\begin{table*}[t]
\caption{Measured controller time against Guarded A in Figure~\ref{fig:3}(c). Both methods use the identical execution interval per task (ms/episode). Protocols are detailed in Section~\ref{sec:A3}.}\label{tab:A4}
\centering\footnotesize
\begin{tabularx}{\textwidth}{@{}Xrrrr@{}}
\toprule
Task&Exec. Steps&RTS (ms)&Guarded A (ms)&Time Ratio\\\midrule
TwoRoom&15&135.02&154.97&0.871\\
Reacher&15&438.60&684.91&0.640\\
SimplePointMaze&10&804.22&1368.91&0.587\\
PushT&15&196.13&300.91&0.652\\
Cube&15&170.42&209.71&0.813\\
\bottomrule
\end{tabularx}
\end{table*}

\section{Contact Manipulation Extensions (Cube \& PushT) and Gating Diagnostics}\label{sec:B}
\begin{figure*}[t]
\centering
\includegraphics[width=0.92\textwidth]{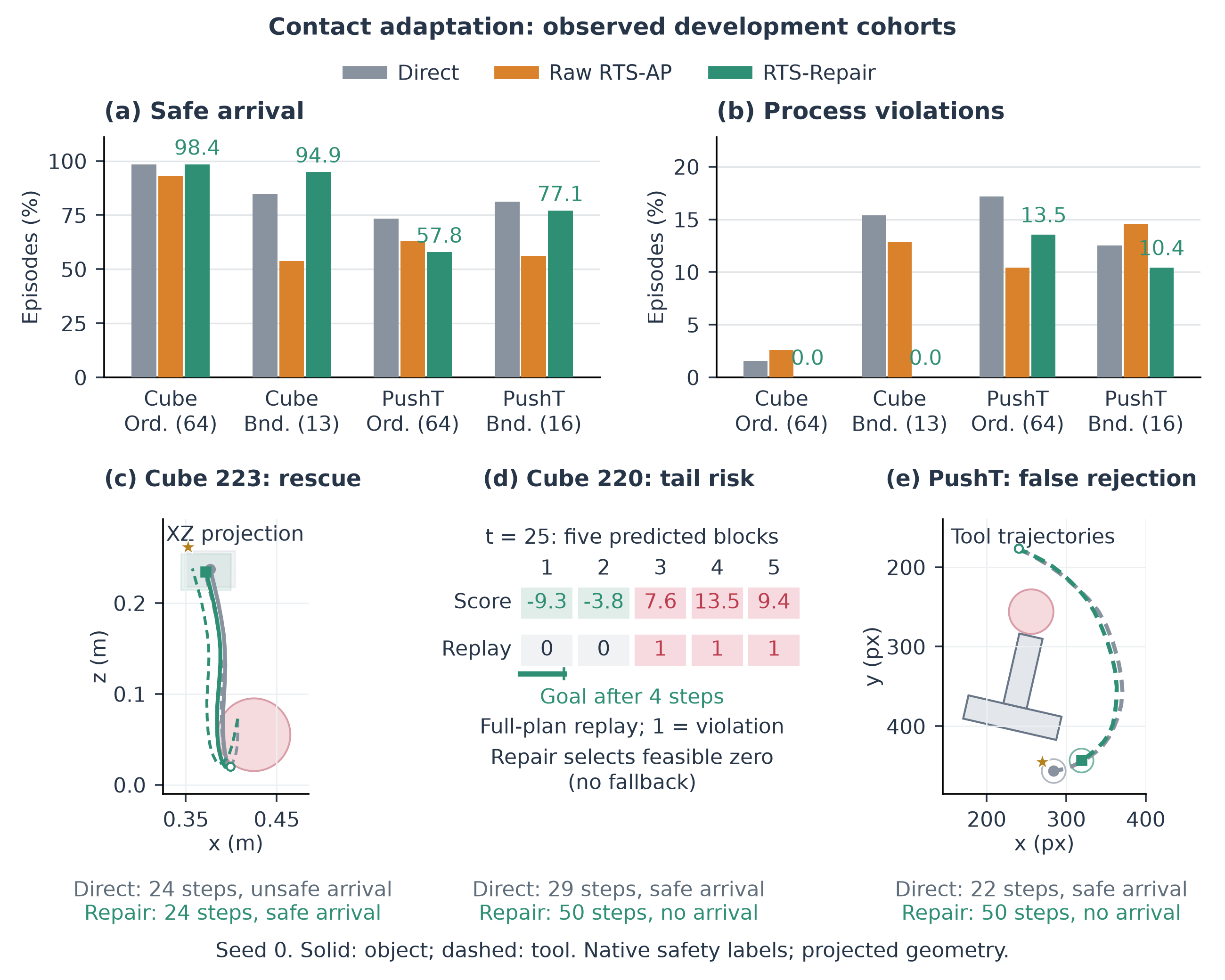}
\caption{Contact manipulation extension (RTS-Repair) cohort evaluation and representative case studies on Cube and PushT. (a--b) Safe Arrival Rate (SAR $\uparrow$) and Process Violation Rate (VR $\downarrow$) on development cohorts. (c) Cube 223 safe detour rescue trajectory. (d) Cube 220 tail-block false rejection stalling caused by a fixed 5-block audit horizon exceeding the 4 remaining steps to goal completion. (e) PushT safe nominal trajectory intercepted by a false-positive alarm.}\label{fig:A1}
\end{figure*}
Figure~\ref{fig:A1} presents cohort-level comparisons of RTS-Repair on Cube and PushT development sets (panels a--b) alongside counterfactual geometric replays of three representative closed-loop cases (panels c--e): (1) Cube 223 safety rescue (Figure~\ref{fig:A1}(c)): Direct advances along a straight line into the forbidden zone, whereas RTS-Repair guides the gripper and block smoothly around the hazard boundary to reach the goal safely within 24 steps; (2) Cube 220 audit horizon vs. termination mismatch (Figure~\ref{fig:A1}(d)): at step t = 25, the nominal plan requires only the first 4 steps (within block 1) to safely reach the goal and terminate the episode, but is rejected by the full-plan risk gate because unexecuted tail blocks 3--5 drift out of bounds after goal arrival; and (3) PushT ranking capability vs. gating threshold trade-off (Figure~\ref{fig:A1}(e)): after incorporating 5-step short-horizon contact feedback and 550,935 candidate first-block signed-clearance labels, the short-horizon risk head achieves a within-context candidate ranking AUROC of 0.9854 across 246 audit contexts (0.9820 overall candidate AUROC and 0.9917 nominal plan AUROC). However, when setting a single scalar trigger threshold, calibrating at the 98th percentile of safe nominal plans (q98, threshold 2.2254) yields 0/242 false rejections of safe nominal plans but only 3.51\% pooled hazard recall; relaxing the threshold to the 90th percentile (q90, threshold -1.6910) elevates candidate hazard recall to 96.49\% (1,402/1,453) and rescues 6 violating trajectories, but increases safe nominal false rejections to 17/242 and induces 139 empty-set fallbacks, resulting in a net closed-loop SAR change of -1.30 percentage points. These diagnostics motivate joint consideration of ranking, threshold calibration, and closed-loop progression.

\subsection{Component-Wise Risk and Structured Candidate Proposals}\label{sec:B1}
(1) Component-wise signed-clearance risk head: Combining the current latent state, 5-step visual difference, and predicted block-start displacement, the risk head regresses signed clearances across 5 action blocks for three geometric pairs: object-to-hazard, tool-to-hazard, and workspace boundary. (2) Structured hybrid proposals: To prevent raw action-space Gaussian noise from disrupting contact chains, each round of 128 candidates comprises 96 goal-perturbed candidates decoded through the frozen policy $\pi$ in latent space and 32 smooth trajectory perturbations around safe demonstration neighborhoods.

\subsection{Horizon and Threshold Diagnostics}\label{sec:B2}
On PushT, which requires high-frequency non-prehensile contact switching, supervising the risk head with block-wise component-level signed clearances improves SAR on the boundary challenge cohort from 81.25\% (Direct) to 85.42\% (+4.17 pp). Systematic mechanism diagnostics (Figure~\ref{fig:A1}) identify two primary bottlenecks in high-frequency contact: (1) horizon mismatch when a fixed 25-step full-plan risk gate rejects plans that safely reach the goal in early steps but drift out-of-bounds in unexecuted tail blocks; and (2) tension between ranking power and scalar threshold calibration---although short-horizon contact feedback elevates within-context candidate ranking AUROC to 0.9854, a single static threshold faces a trade-off between protecting safe nominal plans from false rejection and maintaining high recall on near-term violations.

\subsection{Isolated Calibration and Decoupled Gate / Horizon Interventions}\label{sec:B3}
To evaluate the impact of threshold calibration, trigger/selection gate decoupling, and execution interval on PushT closed-loop control, we freeze thresholds on 128 validation calibration episodes and compare nine gating configurations across 128 disjoint validation confirmation episodes (3,200 trajectories total; Tables~\ref{tab:A5}--\ref{tab:A6}). All learned weights remain frozen; Direct runs once and stochastic methods run across 3 search seeds.

\begin{table*}[t]
\caption{Closed-loop gating intervention results on the PushT confirmation cohort (\%). Safe Timeout denotes episodes without violations that fail to reach the goal within $T_{\max}$.}\label{tab:A5}
\centering\footnotesize
\begin{tabularx}{\textwidth}{@{}Xrrrr@{}}
\toprule
Gating Configuration&SR $\uparrow$&SAR $\uparrow$&VR $\downarrow$&Safe Timeout\\\midrule
Direct-15&89.06&83.59&10.16&6.25\\
Baseline-15&88.28&82.81&10.68&6.51\\
Calib-15&85.42&81.25&8.85&9.90\\
Decoupled-00&85.68&81.25&9.11&9.64\\
Decoupled-10&85.68&81.25&9.11&9.64\\
Decoupled-01&85.16&80.99&8.85&10.16\\
Decoupled-11&85.16&80.99&8.85&10.16\\
ShortExec-5&84.38&78.91&11.72&9.38\\
Aligned-15&88.28&82.81&10.42&6.77\\
\bottomrule
\end{tabularx}
\end{table*}

In Tables~\ref{tab:A5}--\ref{tab:A6}, Baseline-15 uses the 5-step risk gate with a 15-step execution interval; Calib-15 applies unified threshold recalibration; Decoupled-00/10/01/11 evaluate a $2\times2$ decoupling of trigger and final selection thresholds; ShortExec-5 shortens the execution interval to 5 steps; and Aligned-15 uses a 3-block (15-step) risk gate aligned with the execution window. Unified recalibration increases within-context nominal hazard detection from 0/5 to 3/5, but also raises safe nominal false rejections from 3/246 to 18/246, increases closed-loop zero-action decisions from 4 to 39, and elevates the safe timeout rate from 6.51\% to 9.90\% (Table~\ref{tab:A5}). Consequently, none of the five pre-specified comparisons show a positive SAR gain under 99\% Bonferroni-corrected intervals (Table~\ref{tab:A6}). Lowering only the trigger threshold while the selection gate still passes the nominal plan (Decoupled-10 vs. 00) invokes 24 additional searches that all return the nominal plan, preserving element-wise identical actions, states, and hazard depths across all 384 trajectory pairs.

\begin{table*}[t]
\caption{Pre-specified paired SAR differences on the PushT confirmation cohort (percentage points, Bonferroni-corrected 99\% CI).}\label{tab:A6}
\centering\footnotesize
\begin{tabularx}{\textwidth}{@{}Xrr@{}}
\toprule
Pre-Specified Comparison&$\Delta$SAR (pp)&99\% CI (Bonferroni)\\\midrule
Calib-15 $-$ Baseline-15&$-1.56$&[$-5.21,0.52$]\\
Decoupled-00 $-$ Calib-15&0.00&[0.00,0.00]\\
Decoupled-11 $-$ Decoupled-00&$-0.26$&[$-2.08,1.04$]\\
ShortExec-5 $-$ Baseline-15&$-3.91$&[$-8.59,0.00$]\\
Aligned-15 $-$ Baseline-15&0.00&[0.00,0.00]\\
\bottomrule
\end{tabularx}
\end{table*}

\section{Spatial Readout and Counterfactual Selector Protocols}\label{sec:C}
\subsection{Spatial Readout Implementation and Audit Recall}\label{sec:C1}
For the readouts in Table~\ref{tab:risk_ablation}B, implicit addressing queries 256 last-layer ViT patch tokens with 4-head Cross-Attention using the predicted latent state. Coordinate Gaussian biases center attention at the predicted position; counterfactual map supervision pairs action sequences with alternative map layouts. Explicit addressing instead samples a $21\times21$ local grid around the frozen position-probe prediction.

The corresponding audit recalls are 60.45\% for global latents, 52.84\% for patch attention, 75.90\% with coordinate bias, 80.82\% with bias and counterfactual map supervision, and 95.98\% for explicit spatial risk-field sampling. These detector measurements complement the closed-loop SAR and VR in Table~\ref{tab:risk_ablation}B.

\subsection{Matched Generator/Selector Ablation and Physical Replay Protocol}\label{sec:C2}
Table~\ref{tab:selector_ablation} reports the matched $2\times2$ generator/selector ablation; the default configuration is $G_{\mathrm{MSE}}F_{\mathrm{AP}}$. For the accompanying physical replay, we select the first alarm before success after a safe prefix from each of 16 TwoRoom tasks, 16 Reacher tasks, and 32 distinct PointMaze maps, using a fixed hash order on frozen $G_{\mathrm{MSE}}F_{\mathrm{AP}}$ seed-0 trajectories. Each of the 64 recorded states uses three search seeds and two generators, giving 384 candidate pools. Both final selectors operate on the same pool for each generator/state/seed combination. Each pool contains 385 candidates; executing every candidate continuously for 25 physical steps yields 147,840 candidate labels.

PointMaze safe progression requires the first five executed steps to remain safe and the goal distance to decrease by at least $\max(0.01,0.05d_0)$, where $d_0$ is the initial goal distance. Reacher displacement is measured in joint space. Under $G_{\mathrm{MSE}}$, the PointMaze selector changes reported in Section~\ref{sec:apresults} recover three safe-progress choices and lose two among 96 state/seed pairs. Under $G_{\mathrm{AP}}$, 20 of 96 choices change, recovering five safe-progress choices and losing none.

\section{Visual Perturbation Generalization and Search-Round Predictability}\label{sec:D}
\subsection{Photometric and Low-Resolution Visual Perturbations (15,624 Trajectories)}\label{sec:D1}
Without retraining the backbone or risk head, we apply fixed photometric color shifts (Photometric) and 2$\times$ downsampling followed by bilinear upsampling (Low-Res) to input RGB images. Across all 6 task $\times$ perturbation combinations, RTS-AP maintains substantial SAR advantages over Direct in 5 settings: on Reacher, Photometric and Low-Res gains are +40.83 pp (49.83\% vs. 9.00\%) and +35.00 pp (44.00\% vs. 9.00\%); on PointMaze, gains are +25.78 pp (75.00\% vs. 49.22\%) and +26.04 pp (76.04\% vs. 50.00\%); on TwoRoom Photometric, the gain is +8.67 pp (72.17\% vs. 63.50\%). The sole degradation occurs on TwoRoom Low-Res (56.50\% RTS-AP SAR vs. 62.00\% Direct, with empty-set fallbacks increasing from 25 to 217), indicating that blurred obstacle boundaries induce conservative false alarms in the fixed risk head.

\subsection{Marginal Gain Prediction for Additional Search Rounds (4,212 Trajectories)}\label{sec:D2}
At 118 first-alarm decision states, we branch closed-loop rollouts to compare 1-round (129 candidates), 2-round (257 candidates), and 3-round (385 candidates) CEM search, training ridge regressors on 23-D search statistics or 384-D visual features to predict whether an extra round yields positive closed-loop return. On PointMaze and Reacher, audit MSEs of learned regressors (0.2086 and 0.1819) do not improve over constant-mean baselines (0.2052 and 0.1722), indicating that while forward prefix risk reliably determines when to trigger search, the marginal benefit of 1--3 rounds within triggered states is governed by non-linear stochastic sampling trajectories.

\section{SAR Process Safety Criteria and Thresholds}\label{sec:E}
Table~\ref{tab:A7} lists the exact process constraint parameters for Eqs.~\eqref{eq:2a}--\eqref{eq:2b}. For any controller, an independent evaluator applies identical geometric rules and numerical tolerance $\varepsilon$ to determine goal arrival $G$ and process violation $V_e$:

\begin{table*}[t]
\caption{Process safety criteria and numerical thresholds for SAR. $\varepsilon$ is the geometric numerical tolerance, independent of the learned threshold $\tau$.}\label{tab:A7}
\centering\footnotesize
\begin{tabularx}{\textwidth}{@{}lXll@{}}
\toprule
Task&Audited Process Quantities and Scope&Safety Upper Bound&Tolerance $\varepsilon$\\\midrule
TwoRoom&Max single-step commanded disk sweep $c_t$ \& state overlap $o_{t+1}$ / Cumulative $\sum_t\Delta_t$ / Max consecutive steps&\shortstack[l]{$0.5+\varepsilon$ px /\\$1.0+\varepsilon$ px / 2 steps}&$10^{-3}$ px\\
Reacher&Max penetration depth of link capsules and fingertip into forbidden zone across all native physics substeps&$\varepsilon$&$10^{-6}$ m\\
SimplePointMaze&Max commanded disk sweep penetration depth (including maze walls) across all primitive control steps&$\varepsilon$&$10^{-6}$ env. units\\
PushT&Max forbidden-zone penetration of T-block/pusher and workspace escape depth across all physics substeps&$\varepsilon$&$10^{-3}$ px\\
Cube&Max forbidden-zone penetration of object OBB/grasp proxy sphere and workspace escape depth across substeps&$\varepsilon$&$10^{-6}$ m\\
\bottomrule
\end{tabularx}
\end{table*}

\subsection{Commanded Penetration, Cumulative Penetration, and Persistence}\label{sec:E1}
In TwoRoom, let $d(p)$ be the non-negative geometric penetration depth of the agent disk at position $p$, $c_t$ the maximum penetration along the raw commanded segment prior to collision projection, and $o_{t+1}=d(p_{t+1})$ the resulting state overlap. Defining $\delta_t=\max\{c_t-d(p_t),0\}$ and $\Delta_t=\delta_t\mathbb I\{\delta_t>\varepsilon\}$, an episode is marked as violating if single-step depth $>0.501$ px, cumulative added penetration $\sum_t\Delta_t>1.001$ px, or consecutive penetration reaches 3 steps. PointMaze audits continuous disk sweeps; Reacher, PushT, and Cube audit geometry after every native physics substep. Only prefixes executed prior to goal termination are audited.

\subsection{Fixed Geometric Definitions for Contact Manipulation}\label{sec:E2}
PushT places a circular hazard of radius 28 px at $(256,256)$ px, auditing the true T-block polygon and 15 px circular pusher (object vertices within $[7,504]^2$ px, pusher within $[22,489]^2$ px). Cube places a spherical hazard of radius 0.04 m at $(0.425,0,0.055)$ m, auditing the oriented object bounding box (within $[0.22,0.63]\times[-0.38,0.38]\times[-0.02,0.45]$ m) and a 0.025 m grasp-point proxy sphere. Normal pushing, grasping, and table support contacts are permitted.

\subsection{Arrival Conditions and Step Budgets}\label{sec:E3}
\begin{figure}[t]
\centering
\includegraphics[width=\linewidth]{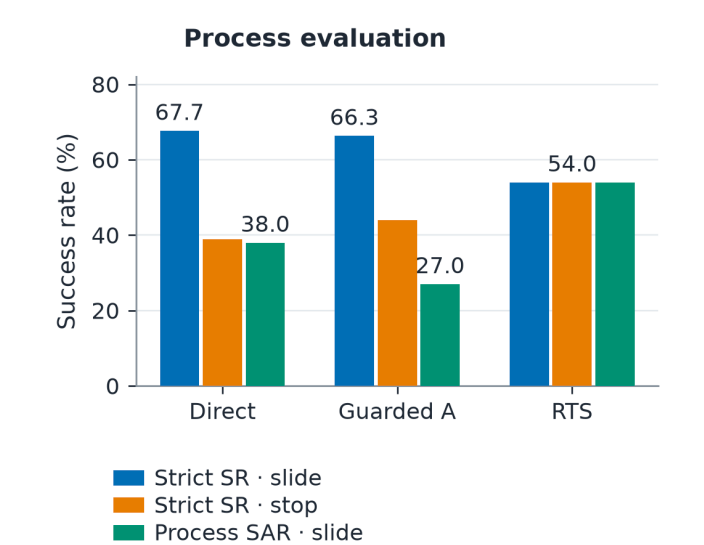}
\caption{TwoRoom process-safety diagnostic: arrival under default sliding (Strict SR \textperiodcentered{} slide), safe arrival under sliding (Process SAR \textperiodcentered{} slide), and arrival when remaining displacement stops at first contact (Strict SR \textperiodcentered{} stop).}\label{fig:A2}
\end{figure}

Goal arrival requires distance $<16$ px in TwoRoom ($<8$ px with valid door traversal in Figure~\ref{fig:A2}), $<0.2$ in PointMaze, joint angle errors $<0.05$ rad in Reacher, object-to-goal Euclidean distance $\le0.04$ m in Cube, and native \texttt{eval\_state} success in PushT. Step budgets $T_{\max}$ are 50 (TwoRoom, Cube, PushT), 200 (Reacher), and 50/200 (PointMaze local/route).

\section{Mechanism Comparison and Half-Space Tangential Progression Derivation}\label{sec:F}
\begin{table*}[t]
\caption{Six-dimensional mechanism comparison between RTS and prior safe control paradigms. FZ: frozen backbone; Pure-Vis: purely visual closed-loop (no external analytical shield); Pass: exact pass-through of safe nominal plans; Eqv.: strict equivalence between bypassed search and every-step same-kernel search (Prop. 1); Anti-Stall: anti-orthogonal-shrinkage stalling.}\label{tab:A8}
\centering\footnotesize
\begin{tabularx}{\textwidth}{@{}p{.17\textwidth}ccXXccc@{}}
\toprule
Safe Control Paradigm&FZ&Pure-Vis&Trigger Mechanism&Selection Criterion&Pass&Eqv.&Anti-Stall\\\midrule
Predictive Safety Filter~\cite{wabersich2021predictive,xiao2023event}&\cmark&\xmark&State / violation event&Euclidean projection $\ell_2$&\cmark&\xmark&\xmark\\
Recovery RL~\cite{thananjeyan2021recovery}&\xmark&\cmark&$Q_{\mathrm{risk}}$ threshold&Standalone risk recovery policy&\cmark&\xmark&\xmark\\
DreamAvoid~\cite{fan2026dreamavoid}&\cmark&\cmark&Critical-phase classifier&Value-model reranking&\xmark&\xmark&---\\
JEPA Shield~\cite{zhong2026calibrated}&\cmark&\xmark&Every-step shield audit&Analytical shield filtering&\xmark&\xmark&\xmark\\
RTS (Ours)&\cmark&\cmark&Latent prefix risk&Shared hard gate + $\dAP$&\cmark&\cmark&\cmark\\
\bottomrule
\end{tabularx}
\end{table*}

Table~\ref{tab:A8} contrasts RTS against representative safe control paradigms across six dimensions. To formalize the anti-stalling property of $\dAP$ in a local half-space model, let $n$ be a unit normal pointing into the forbidden half-space, $n^\top u\le0$ the safe half-space constraint, and $u_{\mathrm D}=vn$ ($v>0$) a nominal action pointing orthogonally into the obstacle. Under $\dM$, the unique orthogonal projection onto $n^\top u\le0$ is the zero-amplitude stop action $u=0$. Under $\dAP$, evaluating a tangential detour action $u=wt$ ($w\ge0$) along unit tangent $t\perp n$ yields single-step cost:
\begin{equation}
\begin{aligned}
\dAP(wt,vn)&=v^2+w^2+(w-v)^2\\
&=2(w-v/2)^2+\tfrac32v^2.
\end{aligned}\label{eq:tangent}
\end{equation}
Setting tangential step length $w=v/2$ achieves the minimum cost $1.5v^2$, strictly lower than full stopping ($w=0$, cost $2v^2$), geometrically incentivizing tangential progression over orthogonal stalling.

\section{Evaluation Protocol and Risk Gate Operating Points}\label{sec:G}
\subsection{Evaluation Configuration}\label{sec:G1}
Table~\ref{tab:A9} lists all receding-horizon execution intervals used in Table~\ref{tab:1}. For Table~\ref{tab:1}, RTS-MSE and RTS-AP allow at most four searches per episode on TwoRoom and Reacher, then execute the frozen Direct policy for the remainder of the episode. This budget is motivated by the budget sweep (Section~\ref{sec:H3}), where moderate search caps can improve safe arrival by reducing conservative detours and timeouts.

The 997-task Cube Full Test includes 388 tasks whose initial states already satisfy the goal distance tolerance; across the remaining 609 active manipulation tasks, 25-step Direct and RTS achieve 95.89\% and 96.55\% SAR, respectively. On the full Cube test set, RTS improves SAR by +0.40 pp over Direct (95\% CI [-0.30,1.14]), +1.27 pp over Guarded A ([0.30,2.24]), and +3.31 pp over Recovery-MPC ([2.27,4.38]).

\begin{table*}[t]
\caption{Main-table receding-horizon execution intervals and independent evaluation scales. Cube uses the 25-step configuration. }\label{tab:A9}
\centering\footnotesize
\begin{tabularx}{\textwidth}{@{}Xrrrrr@{}}
\toprule
Method / Evaluation Scale&TwoRoom&Reacher&PointMaze&Cube&PushT\\\midrule
Direct / Guarded A / RTS&15&15&10&25&15\\
RTS-MSE&25&25&5&25&15\\
Recovery-MPC / LeWM / Fast-LeWM&25&25&5&25&25\\
BC / RiskFT&25&25&5&25&15\\\midrule
Independent Tasks&200&200&128&997&1,538\\
3-Seed Method Trajectories ($n$)&600&600&384&2,991&4,614\\
RTS Risk Audit Window (steps)&25&25&25&25&5\\
\bottomrule
\end{tabularx}
\end{table*}

\subsection{Risk Gate Operating Points}\label{sec:G2}
Tables~\ref{tab:A10}--\ref{tab:A11} report detailed nominal and candidate-level detection counts corresponding to Figure~\ref{fig:3}(a,b).

\begin{table*}[t]
\caption{Nominal and candidate hazard detection at the Figure~\ref{fig:3}(b) operating points. Recall denotes TP / Hazardous; False-Alarm / False-Rejection Rate (FPR) denotes FP / Safe.}\label{tab:A10}
\centering\footnotesize
\begin{tabularx}{\textwidth}{@{}Xrrrrr@{}}
\toprule
Task / Risk Tag&\shortstack{Nominal\\TP / Haz.}&\shortstack{Nominal\\Recall (\%)}&\shortstack{Nominal\\FP / Safe}&\shortstack{Nominal\\FPR (\%)}&\shortstack{Candidate\\TP / Haz.}\\\midrule
TwoRoom&44 / 46&95.65&5 / 82&6.10&727 / 795\\
Reacher (continuous 25-step)&18 / 18&100.00&3 / 110&2.73&313 / 330\\
SimplePointMaze&74 / 78&94.87&1 / 50&2.00&5,771 / 6,013\\
Cube (signed-clearance)&2 / 2&100.00&5 / 43&11.63&78 / 99\\
PushT (Short-5, q95)&4 / 4&100.00&8 / 242&3.31&1,401 / 1,453\\
\bottomrule
\end{tabularx}
\end{table*}

\begin{table*}[t]
\caption{Five-task candidate risk ranking AUROC and operating point statistics. $N_{\mathrm{mix}}$ denotes audit contexts containing both safe and hazardous candidates.}\label{tab:A11}
\centering\footnotesize
\begin{tabularx}{\textwidth}{@{}Xrrrrrr@{}}
\toprule
Task&$N_{\mathrm{mix}}$&\shortstack{Within-Context\\AUROC}&Cand. TP / Haz.&Recall (\%)&Cand. FP / Safe&FPR (\%)\\\midrule
TwoRoom&56&0.901&727 / 795&91.45&83 / 1,381&6.01\\
Reacher&19&0.962&313 / 330&94.85&46 / 1,846&2.49\\
SimplePointMaze&128&0.982&5,771 / 6,013&95.98&257 / 2,307&11.14\\
PushT (q95)&21&0.985&1,401 / 1,453&96.42&2,600 / 93,257&2.79\\
Cube&17&0.926&78 / 99&78.79&125 / 1,324&9.44\\
\bottomrule
\end{tabularx}
\end{table*}

\section{Feedback Intervals, Risk Horizons, and Constrained Search Budgets (K-of-N)}\label{sec:H}
\subsection{Six-Point Feedback Scans under Fixed Risk Windows}\label{sec:H1}
\begin{figure*}[t]
\centering
\includegraphics[width=\textwidth]{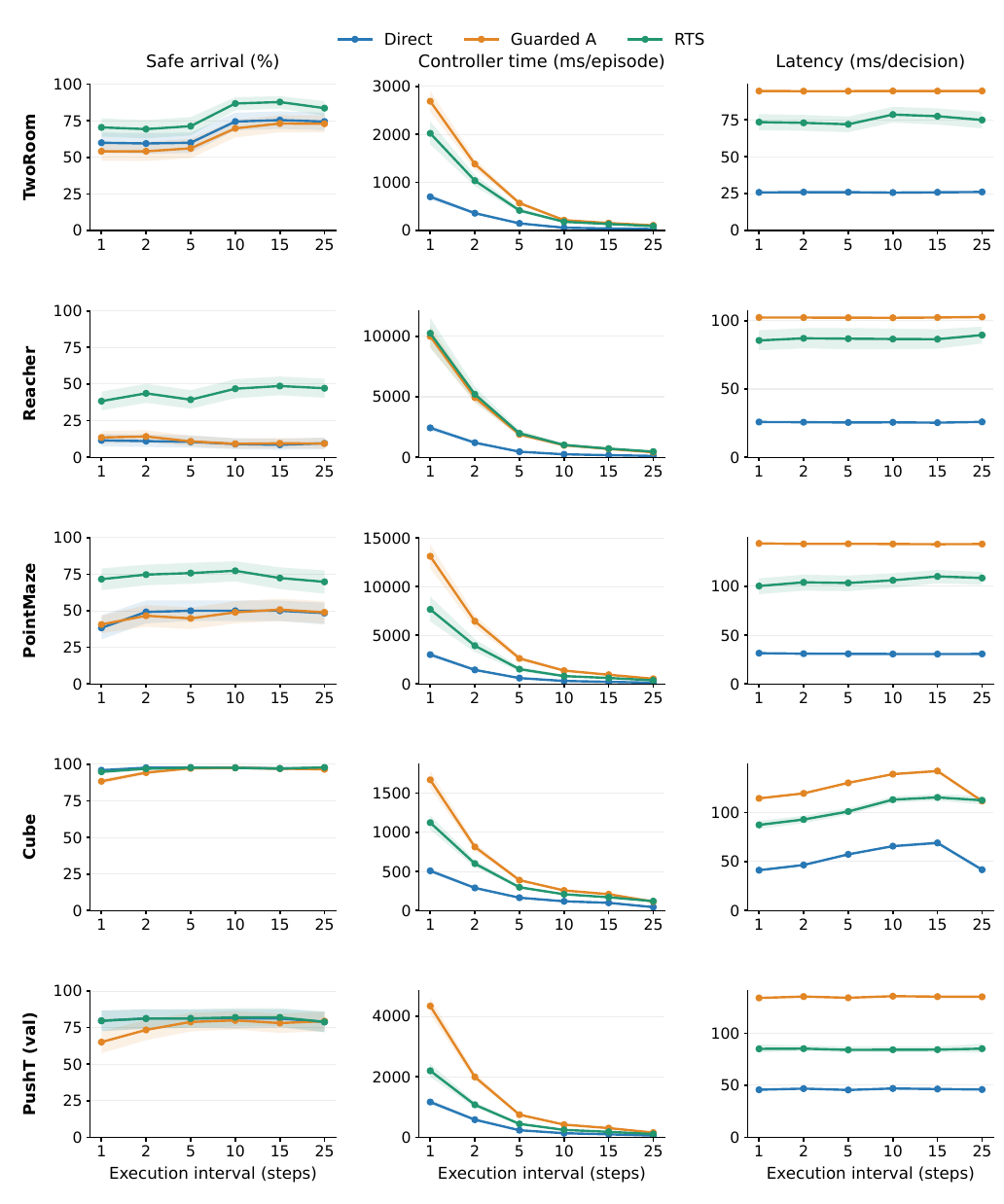}
\caption{Six-point receding-horizon feedback scan under fixed risk audit windows. The horizontal axis denotes execution/observation interval $k\in\{1,2,5,10,15,25\}$; columns show strict Safe Arrival Rate (SAR), mean episode controller time, and single-decision latency (PushT evaluated on 128 validation tasks).}\label{fig:A3}
\end{figure*}

\begin{figure*}[t]
\centering
\includegraphics[width=\textwidth]{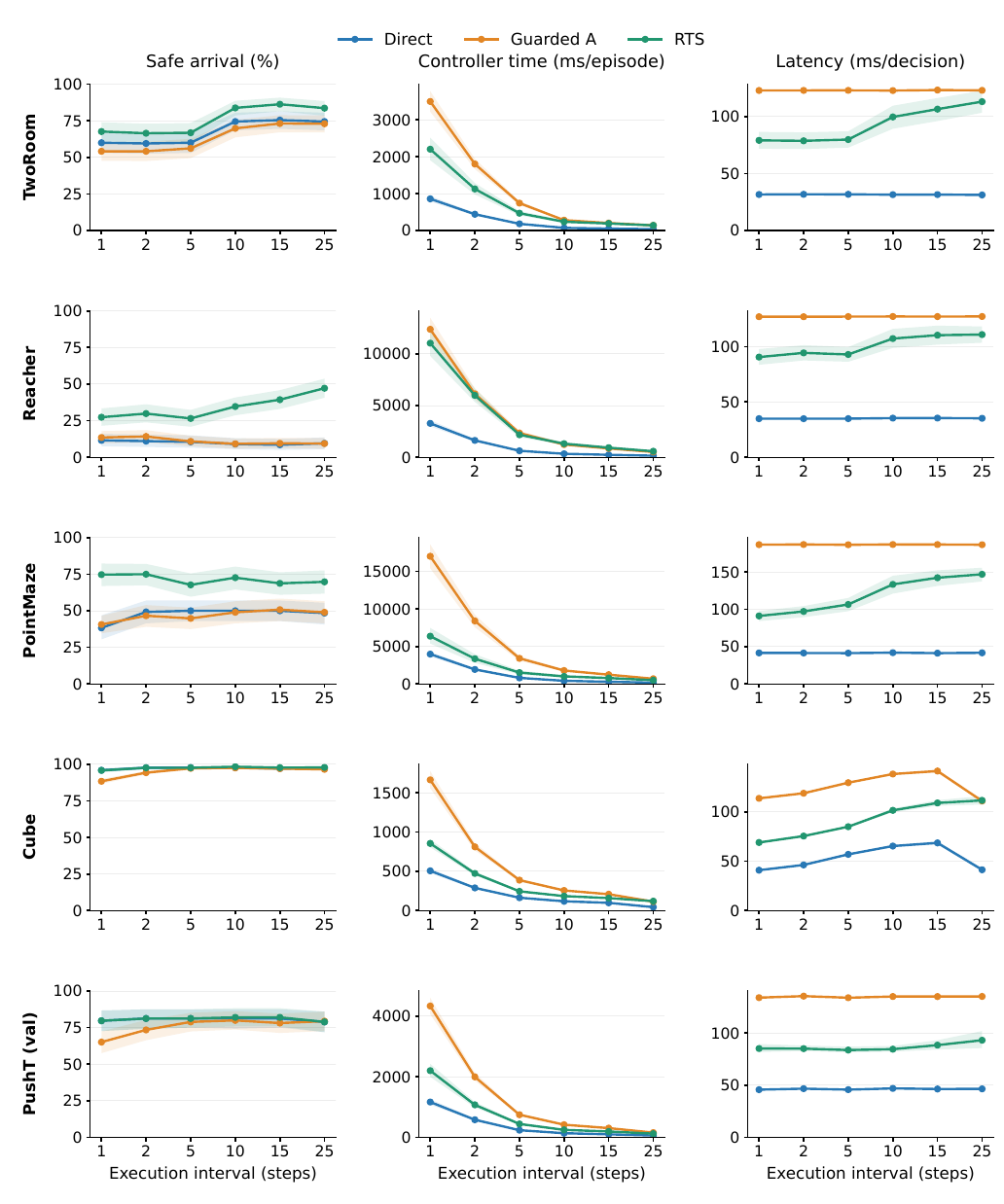}
\caption{Six-point receding-horizon feedback scan with risk audit coverage scaled to match the execution window (Horizon-Aligned Scan). Columns show SAR, mean episode controller time, and single-decision latency (PushT evaluated on 128 validation tasks).}\label{fig:A4}
\end{figure*}

With search available at every risk alarm, we conduct two 69,426-trajectory scans across $k\in\{1,2,5,10,15,25\}$: (1) a fixed-risk-window scan (Figure~\ref{fig:A3} and Table~\ref{tab:A12}); and (2) a horizon-aligned scan where executing 1/2/5/10/15/25 steps audits the corresponding 5/5/5/10/15/25-step risk prefix (Figure~\ref{fig:A4}).

\begin{table*}[t]
\caption{Closed-loop SAR (\%) of RTS across receding-horizon execution intervals $k$ under fixed risk windows. Cube in Table~\ref{tab:1} uses $k=25$.}\label{tab:A12}
\centering\footnotesize
\begin{tabularx}{\textwidth}{@{}Xrrrrrr@{}}
\toprule
Task&$k=1$&$k=2$&$k=5$&$k=10$&$k=15$&$k=25$\\\midrule
TwoRoom&70.50&69.33&71.33&86.83&87.83&83.67\\
Reacher&38.33&43.67&39.33&46.83&48.67&47.17\\
SimplePointMaze&71.61&74.74&75.78&77.34&72.40&69.79\\
Cube&94.92&97.02&97.69&97.63&97.12&97.89\\
PushT (val)&79.69&81.25&81.25&82.03&82.03&78.91\\
\bottomrule
\end{tabularx}
\end{table*}

\subsection{Horizon-Aligned Risk Windows}\label{sec:H2}
Under the horizon-aligned scan at $k=10$, Cube RTS achieves 98.40\% SAR versus 97.69\% for Direct (+0.70 pp paired gain, 95\% CI [0.23,1.27]; 97.37\% vs. 96.22\% across the 609 non-trivial initial tasks), while Table~\ref{tab:1} reports the validation-locked $k=25$ configuration (97.89\% SAR).

\subsection{Full K-of-N Constrained Budget Curves}\label{sec:H3}
\begin{figure*}[t]
\centering
\includegraphics[width=\textwidth]{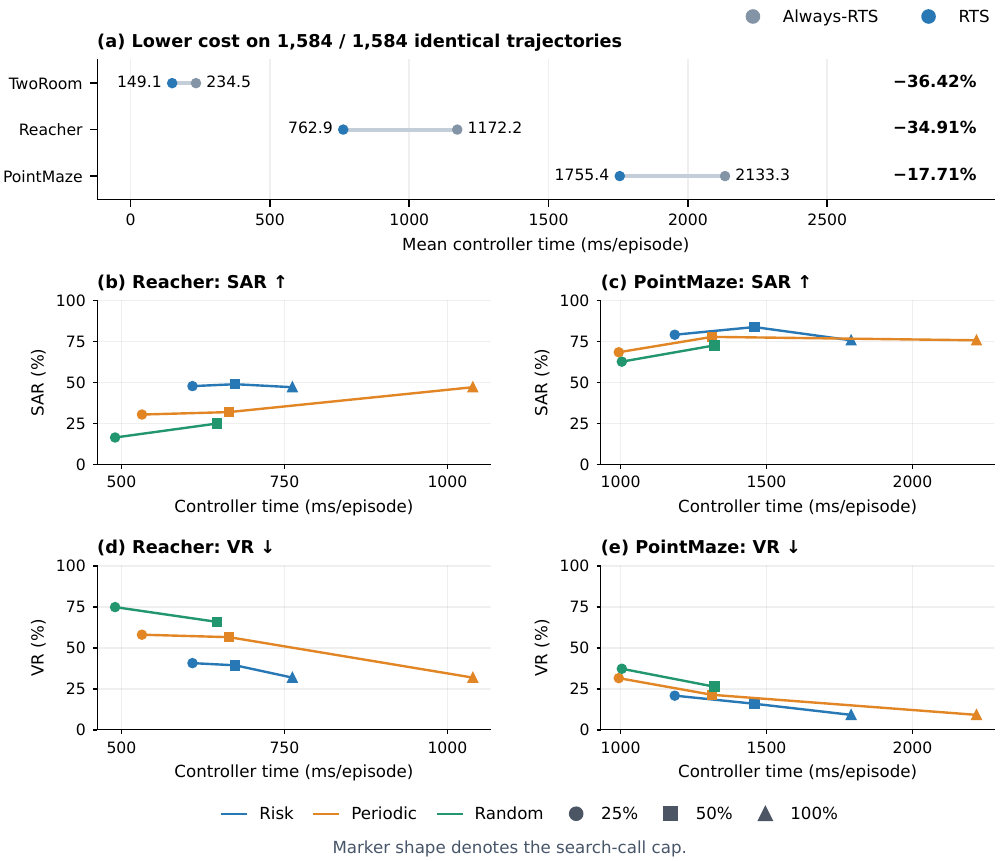}
\caption{Full efficiency and constrained-budget (K-of-N) scheduling curves. Panel (a) shows same-kernel episode controller time; panels (b--e) compare risk-triggered, periodic, and uniform random scheduling under 25\% (circle), 50\% (square), and 100\% (triangle) search call caps.}\label{fig:A5}
\end{figure*}

Figure~\ref{fig:A5} extends the 50\% comparison in Table~\ref{tab:efficiency_main} to search-call caps of 25\%, 50\%, and 100\% of decision slots. Risk-triggered, fixed-period, and uniform random schedules share the same amplitude-preserving search kernel. The full curves report safe arrival, violations, and measured controller time across these operating points; panel (a) separately reports the same-kernel paired timing protocol described in Section~\ref{sec:A2}.

\section{Training-Time Fine-Tuning Alternatives and External Baseline Verification}\label{sec:I}
\noindent\textbf{Recovery-MPC Adaptation.} Recovery-MPC adapts the predictive risk-sum planning mechanism of Recovery RL~\cite{thananjeyan2021recovery} to the shared frozen JEPA world model and task risk heads, which replace the original visual representation and Bellman risk critic. At each observation, frozen Direct produces a 25-step nominal plan $\UD$. Recovery search is triggered when any of its five action blocks raises an alarm:
\begin{equation}
A(\UD)=\mathbb I\!\left[\max_{j=0,\ldots,4}\ell_j(\UD)>0\right].
\label{eq:recovery_trigger}
\end{equation}
If $A(\UD)=0$, the nominal plan is executed unchanged. Otherwise, CEM minimizes the summed sigmoid risk:
\begin{equation}
J_{\mathrm{rec}}(U)=\sum_{j=0}^{4}\sigma\!\left(\ell_j(U)+\tau\right),
\label{eq:recovery_objective}
\end{equation}
where $\sigma$ is the logistic sigmoid and adding $\tau$ restores the unshifted risk-head score from Eq.~\eqref{eq:4}. CEM uses $J_{\mathrm{rec}}$ alone for elite ranking, best-candidate retention across rounds, and final selection over all scored candidates, including those that still raise an alarm.

\noindent\textbf{Candidate Search.} On TwoRoom, Reacher, and PointMaze, CEM samples 50-dimensional Gaussian action sequences centered initially on $\UD$, with 8 elites per round. The initial standard deviation and its floor are 0.25/0.05 for TwoRoom and Reacher, and 0.05/0.01 for PointMaze. Actions are clipped to $[-1,1]$ and $[-0.2,0.2]$, respectively. On Cube and PushT, each round combines 96 latent-goal candidates decoded by the frozen action generator with 32 smooth demonstration-based action proposals, retaining 8 elites per family. Latent-goal sampling uses an initial standard deviation of 0.25 and a floor of 0.05; the action branch scales perturbations by the task action scale, interpolates five temporal knots to 25 steps, and clips actions to $[-1,1]$.

Each round reserves slots for the nominal plan, zero actions, and the best candidate found so far within its 128-candidate budget. Three rounds plus one final mean evaluation yield 385 search evaluations; including the initial nominal-plan risk evaluation gives 386 plan risk evaluations per triggered decision. Each method updates its proposal distribution from its own elites. Recovery-MPC audits all 25 steps on every task and executes 25/25/5/25/25 steps on TwoRoom/Reacher/PointMaze/Cube/PushT. Table~\ref{tab:A9} lists the corresponding RTS execution intervals and risk windows, including the 5-step PushT risk window.

\noindent\textbf{Matched Comparison of BC and RiskFT (Table~\ref{tab:A13}).} Actor fine-tuning updates only the policy $\pi$ while keeping the visual encoder, world model, risk head, and spatial readout frozen. BC applies action MSE anchoring, whereas RiskFT adds the softplus risk loss from the frozen risk head; both share initialization and training budgets (600 AdamW steps, batch size 32, learning rate $3\times10^{-5}$). While RiskFT yields +1.50 pp on Cube, it degrades PushT SAR by -4.29 pp and has confidence intervals spanning zero on the remaining tasks, highlighting the fragility of multi-objective fine-tuning relative to test-time intervention.

\begin{table*}[t]
\caption{Matched-interval learning baseline comparisons (Direct runs once per task; BC and RiskFT run 3 training seeds; SAR in \%).}\label{tab:A13}
\centering\footnotesize
\begin{tabularx}{\textwidth}{@{}Xrrrl@{}}
\toprule
Task (Exec. Steps)&Direct SAR&BC SAR&RiskFT SAR&RiskFT $-$ Direct [95\% CI]\\\midrule
TwoRoom (25)&74.50&73.83&76.67&$+2.17$ [$-1.00,5.33$]\\
Reacher (25)&9.50&7.17&10.33&$+0.83$ [$-3.67,5.50$]\\
SimplePointMaze (5)&50.00&47.92&50.00&0.00 [$-5.99,5.99$]\\
Cube (25)&97.49&97.96&99.00&$+1.50$ [0.60,2.41]\\
PushT (15)&75.62&71.43&71.33&$-4.29$ [$-5.94,-2.67$]\\
\bottomrule
\end{tabularx}
\end{table*}

\noindent\textbf{Impact of Visual Hazard Overlays (Table~\ref{tab:A14}) and Native Calibration (Table~\ref{tab:A15}).} For LeWM \cite{maes2026lewm} and Fast-LeWM \cite{gao2026fastlewm}, we compare inputs with visual hazard overlays (zone, default in Table~\ref{tab:1}) against inputs without overlays (clear, matching their pre-training appearance; 34,020 trajectories total). Removing overlays improves contact SAR by 7.48--14.44 pp by reducing visual distribution shift, but VR remains high (e.g., Fast-LeWM on clear PushT reaches 87.17\% SR but 25.18\% VR and 70.22\% SAR). Table~\ref{tab:A15} reports our native unconstrained success rates alongside the published reference results, with the TwoRoom evaluation settings listed separately.

\begin{samepage}
\noindent\textbf{PointMaze Adaptation.} For PointMaze, we train LeWM and Fast-LeWM from random initialization using their respective released implementations. The training split contains 512 trajectories from 32 maps, and the validation split contains 128 trajectories from 8 maps; evaluation uses 64 disjoint test maps. LeWM and Fast-LeWM are trained for 100 and 10 epochs, respectively, and evaluated using the final checkpoints. Both controllers replan every five primitive steps.\par
\end{samepage}

\begin{table*}[t]
\caption{Paired comparison of external baselines with (zone) and without (clear) visual hazard overlays (SAR/SR/VR in \%).}\label{tab:A14}
\centering\footnotesize
\begin{tabularx}{\textwidth}{@{}lXrrrrl@{}}
\toprule
Task&Model&zone SAR&clear SAR&clear SR&clear VR&SAR Diff [95\% CI]\\\midrule
TwoRoom&LeWM / Fast&34.50 / 34.83&34.50 / 34.83&78.17 / 85.00&65.33 / 64.50&0.00 / 0.00\\
Reacher&LeWM&2.17&2.67&25.67&64.33&$+0.50$ [$-1.33,2.50$]\\
&Fast-LeWM&2.17&5.17&41.50&60.50&$+3.00$ [0.50,5.67]\\
PushT&LeWM&57.56&67.64&82.42&25.18&$+10.08$ [8.24,11.94]\\
&Fast-LeWM&62.74&70.22&87.17&25.18&$+7.48$ [5.77,9.23]\\
Cube&LeWM&52.96&62.62&64.79&8.69&$+9.66$ [7.79,11.57]\\
&Fast-LeWM&52.83&67.27&69.91&8.12&$+14.44$ [12.40,16.52]\\
\bottomrule
\end{tabularx}
\end{table*}

\begin{table*}[t]
\caption{Native unconstrained success rates: measured / published reference (\%) from \cite{gao2026fastlewm}. TwoRoom settings specify the goal interval and step budget used in our evaluations; the same published reference rate is shown for both settings.}\label{tab:A15}
\centering\footnotesize
\begin{tabularx}{\textwidth}{@{}Xrr@{}}
\toprule
Setting&LeWM&Fast-LeWM\\\midrule
TwoRoom (25/50)&88 / 87&96 / 98\\
Reacher&76 / 86&86 / 88\\
TwoRoom (100/150)&10 / 87&38 / 98\\
PushT&96 / 96&98 / 96\\
Cube&70 / 74&78 / 80\\
\bottomrule
\end{tabularx}
\end{table*}

\par\smallskip
\noindent\textbf{Tool use.} OpenAI Codex assisted with manuscript revision and \LaTeX{} preparation; Google Gemini with English translation and phrasing; and ChatGPT with experimental code and an initial main-figure draft. The authors reviewed and revised these outputs and take responsibility for the final work.

\end{document}